%% file: main.tex
\def\year{2021}\relax
\documentclass[letterpaper]{article} 

\usepackage{xifthen}
\usepackage{subcaption}
\usepackage{derivative}
\usepackage{amsthm}
\usepackage{siunitx}
\usepackage{amsmath,amsfonts,amssymb}
\usepackage{mathtools}
\usepackage{thmtools}
\usepackage{graphicx}
\usepackage{siunitx}
\usepackage{booktabs}
\usepackage{cleveref}
\usepackage{adjustbox}
\usepackage{soul}
\usepackage{appendix}
\usepackage{array}
\usepackage{stackengine,wasysym}
\usepackage{array,multirow,makecell}
\usepackage{tabularx}
\usepackage[ruled, linesnumbered, commentsnumbered, longend,noline]{algorithm2e}
\usepackage[switch]{lineno}
\usepackage{comment}

\newcommand{\sysname}{\textsc{HyDRA}}

\input{latex_definition/math}
\input{latex_definition/machinelearning}
\input{latex_definition/theorem}

\input{definitions}

\usepackage{aaai21}  
\usepackage{times}  
\usepackage{helvet} 
\usepackage{courier}  
\usepackage[hyphens]{url}  
\usepackage{graphicx} 
\usepackage{natbib}  
\usepackage{caption} 
\title{\sysname: Hypergradient Data Relevance Analysis for Interpreting\\ Deep Neural Networks}
\author{
  Yuanyuan Chen\textsuperscript{\rm 1},
  Boyang Li\textsuperscript{\rm 1, \rm 2*},
  Han Yu\textsuperscript{\rm 1*},
  Pengcheng Wu\textsuperscript{\rm 1},
  and Chunyan Miao\textsuperscript{\rm 1*}
  \\
}
\affiliations{
  \textsuperscript{\rm 1}School of Computer Science and Engineering, Nanyang Technological University \\
  \textsuperscript{\rm 2}Alibaba-NTU Singapore Joint Research Institute  \\
  \{yuanyuan.chen, boyang.li, han.yu, pengcheng.wu, ascymiao\}@ntu.edu.sg\\
  *Corresponding authors
}

\begin{document}

\maketitle

\begin{abstract}
  The behaviors of deep neural networks (DNNs) are notoriously resistant to human interpretations. In this paper, we propose Hypergradient Data Relevance Analysis, or \sysname{}, which interprets the predictions made by DNNs as effects of their training data. Existing approaches generally estimate data contributions around the final model parameters and ignore how the training data shape the optimization trajectory. By unrolling the hypergradient of test loss w.r.t.\ the weights of training data, \sysname{} assesses the contribution of training data toward test data points throughout the training trajectory. In order to accelerate computation, we remove the Hessian from the calculation and prove that, under moderate conditions, the approximation error is bounded. Corroborating this theoretical claim, empirical results indicate the error is indeed small. In addition, we quantitatively demonstrate that \sysname{} outperforms influence functions in accurately estimating data contribution and detecting noisy data labels. The source code is available at~https://github.com/cyyever/aaai\_hydra.
\end{abstract}

\section{Introduction}

\emph{What makes neural networks do exactly what they do?} This is a crucial question that lingers in the minds of machine learning researchers and practitioners. The underpinnings of deep neural networks (DNNs), including non-convex objective functions, stochastic optimization, and over parameterization, may implicitly regularize the network and improve generalization~(e.g., \citealt{NEURIPS2018_54fe976b}; \citealt{arora2018optimization}), but also hinder the interpretation of the network behaviors and the training process.
The black-box nature of DNNs can render their predictions untrustworthy in the eyes of the general public and prohibit wide adoption. Conversely, good interpretability can enhance the training, debugging, and auditing of DNNs.

The focus of the present paper is to understand DNNs by attributing their predictions to the training data. That is, how training data influence network predictions on test data. However, as the training data are involved in the entirety of the complex training process, accurately capturing their influence on the final network is a challenge.

\begin{figure}[t]
  \centering
  \includegraphics[scale=0.70]{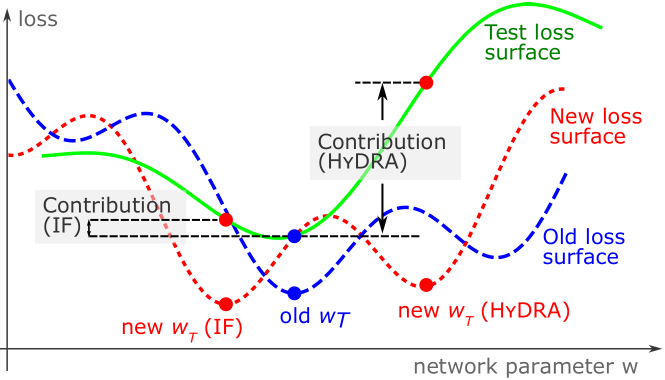}
  \caption{An illustration of \sysname{} and influence functions (IF). After the removal of some training data points, the training loss shifts from the blue curve to the red curve. IF estimates the contribution within the convex region near the old $\param[T]$.~\sysname{} tracks the influence of data removal along the entire optimization process, possibly leading to a different local optimum.}\label{fig:hydra-vs-if}
\end{figure}

The pioneering work of \citep{koh2017understanding} proposes influence functions (IF), which measure the contribution of a training sample $\vect{z}_{i}$ to a test sample $\ztest$ as the change in the loss of $\ztest$ when the weight of training data $\vect{z}_{i}$ is changed marginally.

This technique enables a number of applications (e.g., \citealt{pmlr-v119-alaa20b}), but suffers from two drawbacks. First, it only measures the contribution around the final model parameters $\param[T]$ and ignores the possibility that a change in the weights of training data may lead to a local optimum different from $\param[T]$ with a drastically different test loss.
Second, IF relies on the inverse Hessian on the whole training data, which is computationally expensive to approximate.~\citep{Satoshi2019} analyzed the entire trajectory but still rely on Hessians.

In this paper, we propose Hypergradient for Data Relevance Analysis (\sysname) to address the aforementioned shortcomings. By unrolling the test loss gradient with respect to data weights through all training steps, \sysname{} accounts for how changes in data weights shape the whole training process. As optimization utilizes the gradient on the same data repeatedly, even marginal changes in the data weights can accumulate and eventually shift the point of convergence. Figure~\ref{fig:hydra-vs-if} illustrates such a scenario where the local analysis of IF and the whole-of-trajectory analysis of \sysname{} find separate local minima and contribution values. Further, to simplify computation, we propose an approximation method that eliminates the cumbersome Hessian or its inverse and establishes an analytical upper bound on the approximation error.

Empirically, we verified that the approximation indeed results in small error and is closer to the whole-of-trajectory Hessian-aware measurements than IF.
The approximation method of \sysname{} achieved high correlation with the Hessian-aware measurements, whereas the local analysis of IF added \num{11}\% to \num{16}\% to the approximation error. In the particular experiment we conducted, the removal of Hessian reduces wall-clock running time by a factor of \num{971} on \num{2} TitanX Pascal GPUs.
We also compared \sysname{} and IF in their ability to identify data points with erroneous labels and found that\sysname{} had provided superior detection performance.



\section{\sysname{}: Hypergradient for Data Relevance Analysis}

We first introduce some preliminaries. We aim to learn a function $f_{\param}: \mathcal{X} \rightarrow \mathcal{Y}$ parameterized by $\param$ in the hypothesis space \hypothesisspace. The training dataset contains $N$ data points and is denoted as \(\trainingdataset =\{\ztrain_{i}\}{}_{i=1}^N\), where $\ztrain_{i} \in \mathcal{X} \times \mathcal{Y}$. The function $f_{\param}$ is learned by minimizing both the empirical risk $\empiricalrisk$ and the regularization term $\regularizer(\param)$:
\begin{align}
  \begin{split}
    \label{eq:train-loss}
    \empiricalrisk(\param) & = \sum_{\vect{z}_i \in \trainingdataset} \left(1/N+\epsilon_i \right) \, \persampleloss{i}{}, \\
    \trainingloss(\param)  & = \empiricalrisk(\param) + \weightdecay \regularizer(\param),
  \end{split}
\end{align}
where $\lossfun$ is the per-sample loss such as cross entropy, and $\weightdecay$ is a regularization coefficient. In this paper, we adopt $L^2$ regularization $\regularizer(\param) = \ltworegularizer$.
The sample loss weight $\epsilon_i$ is set to zero during training; its function will be explained shortly.
See table~\ref{tab:notations} for the notations used in this paper.


Similar to the training dataset, the test dataset with $M$ data points is denoted as $\testdataset = \{ \ztest[i]\}{}_{i=1}^M$ with $\trainingdataset \cap \testdataset = \varnothing$.
The test loss \testloss{} is a measurement of model generalizability:
\begin{equation*}
  \testloss(\param[T]) = \frac{1}{M} \sum_{\ztest \in \testdataset} \lossfun(\ztest,\param[T]).
\end{equation*}

We usually optimize the loss function using vanilla Gradient Descent (GD) or GD with momentum.
Vanilla GD iteratively updates the parameters $\param$ as
\begin{equation*}
  \param[t] = \param[t-1] - \learningrate \vect{g}_{t-1},
\end{equation*}
where \learningrate{} is the learning rate at step t. In GD with momentum, we first update the momentum \mom[t], followed by updating \param[t].
\begin{equation*}
  \mom[t] = p\mom[t-1] + \vect{g}_{t-1},
\end{equation*}
\begin{equation*}
  \param[t] = \param[t-1] - \learningrate \mom[t],
\end{equation*} where $p$ is a constant in interval $(0, 1)$ that determines the strength of the accumulated momentum. We repeat the optimization for $T$ steps and arrive at the final network parameters \param[T]. The stochastic versions of the two algorithms simply replace $\vect{g}_{t}$ with the gradients on mini-batches of training data.

\subsection{Measuring Data Contribution}

\begin{table}[t]
  \renewcommand{\arraystretch}{1.2}
  \caption{Frequently used notations}
  \centering
  \begin{tabular}{ll}
    \toprule
    $\trainingdataset$       & Training dataset                                                                                   \\
    $\testdataset$           & Test dataset                                                                                       \\
    $N$                      & Size of the training dataset                                                                       \\
    $\param[t]$              & Model parameters at training step $t$                                                              \\
    $\epsilon_i$             & Marginal weight for the $\ith$ training sample                                                     \\
    $\ztrain_{i}$            & $\ith$ training sample                                                                             \\
    $\ztest[j]$              & $\jth$ test sample                                                                                 \\
    $\regularizer(\param)$   & Regularization term, assuming the form $\ltworegularizer$                                          \\
    $\weightdecay$           & Regularization coefficient                                                                         \\
    $\learningrate[t]$       & Learning rate at training step $t$                                                                 \\
    $p$                      & Momentum factor                                                                                    \\
    $\empiricalrisk(\param)$ & Empirical risk at model parameters $\param$                                                        \\
    $\trainingloss(\param)$  & Training loss at model parameters $\param$                                                         \\
    $\testloss(\param)$      & Test loss at model parameters $\param$                                                             \\
    $\vect{g}_t$             & Model gradient at training step $t$, $\pdv{\trainingloss{}(\param[t])}/{\param}$                   \\
    $\hessianer[t]$          & Empirical risk Hessian at training step $t$, $\pdv[order=2]{\empiricalrisk{}(\param[t])}/{\param}$ \\
    $\hessian[t]$            & Model Hessian at training step $t$, $\pdv[order=2]{\trainingloss{}(\param[t])} /{\param}$          \\
    $\nabla_{t, i}$          & Shorthand for $\odv{\param[t]}/{\epsilon_i}$                                                       \\
  \end{tabular}
  \label{tab:notations}
\end{table}

Recall that every training sample $\ztrain_i$ is associated with the data weight $1/N + \epsilon_i$ in the training loss (equation~\ref{eq:train-loss}). Thus, we can remove the \ith{} sample from the training by setting $\epsilon_i$ to $-1/N$. The resulting change in the test loss can be estimated with a first-order Taylor expansion. We define the contribution of the $\ith{}$ training data point on model performance, $\mathcal{C}(i)$, as the estimated change (which is negligible with a sufficiently large training dataset), as
\begin{equation}
  \label{eq:contribution}
  \mathcal{C}(i) \defeq -\frac{1}{N} \odv{\testloss(\param[T])}{\epsilon_i}\Bigr|_{\epsilon_i=0}.
\end{equation}
If removal of the data sample $\ztrain_i$ from the training data increases test loss, the sample would have a positive contribution. Otherwise, the sample hurts generalization performance and makes a negative contribution.

We can measure the contribution to any portion of the test set simply by replacing $\testloss$ in equation~\eqref{eq:contribution} with the loss on the data points in question. For example, the contribution by the $\ith$ training sample on model performance regarding the $\jth$ test sample, $\mathcal{C}(i, j)$, is computed as
\begin{equation*}
  \begin{split}
    \mathcal{C}(i, j) & \defeq -\frac{1}{N}\odv{\lossfun(\ztest[j], \param[T])}{\epsilon_i}\Bigr|_{\epsilon_i=0}.
  \end{split}
\end{equation*}


In equation~\eqref{eq:contribution}, $\testloss$ does not directly depend on $\epsilon_i$ and only through \param, we can write the total derivative as
\begin{equation}
  \label{eq:val_contrib}
  \odv{\testloss(\param[T])}{\epsilon_i} = \pdv{\testloss(\param[T])}{\param} \odv{\param[T]}{\epsilon_i}.
\end{equation}
$\pdv{\testloss(\param[T])}/{\param}$ can be exactly computed using backpropagation. However, computing $\odv{\param[T]}/{\epsilon_i}$, or in other words $\paramdiff{T}{i}$, is not so straightforward as $\epsilon_i$ is involved in the entire optimization process. In the following, we give the recurrence equations for $\paramdiff{T}{i}$ in the whole-batch gradient descent.

In vanilla gradient descent, $\paramdiff{t}{i}$ can be computed recurrently as a function of $\paramdiff{t-1}{i}$ as follows:
\begin{equation} \label{eq:sgd-hypergradient}
  \begin{split}
    \paramdiff{t}{i} & = \paramdiff{t-1}{i} - \learningrate \odv{\vect{g}_{t-1}}{\epsilon_i}                  \\
                     & = \paramdiff{t-1}{i} - \learningrate \hessianer[t-1] \paramdiff{t-1}{i}                \\
                     & \quad -\learningrate \weightdecay \paramdiff{t-1}{i} - \learningrate \vect{g}_{t-1,i},
  \end{split}
\end{equation}
where $\vect{g}_{t,i}$ denotes the gradient of \ith{} sample.
Note that since $\vect{g}_t$ is a function of both $\param[t]$ and $\epsilon_i$, we have $\odv{\vect{g}_t}{\epsilon_i} = \hessian[t] \paramdiff{t}{i}+\vect{g}_{t-1,i}$, which introduces the Hessian.

Similarly, for gradient descent with momentum,
\begin{align}
  \label{eq:momentum_diff_1}
  \odv{\mom[t]}{\epsilon_i} & = p \odv{\mom[t-1]}{\epsilon_i} + \hessianer[t-1] \paramdiff{t-1}{i} \\
                            & \quad + \weightdecay \paramdiff{t-1}{i}+  \vect{g}_{t-1,i},          \\
  \paramdiff{t}{i}          & = \paramdiff{t-1}{i} - \learningrate \odv{\mom[t]} {\epsilon_i}.
\end{align}
In both cases we have the initial conditions
\begin{align*}
  \paramdiff{0}{i} & = \vect{0} \text{\, and \,} \momdiff{0}{i} = \vect{0}.
\end{align*}
By recurrently computing $\paramdiff{t}{i}$ through the entire optimization trajectory, \sysname{} holistically measures the contribution of $\ith{}$ data point to the neural network.

\begin{algorithm}[t]
  \SetKwInOut{KwIn}{Input}
  \SetKwInOut{KwOut}{Output}
  \SetNoFillComment
  \KwIn{traced training sample $ \ztrain_i$, training dataset size $N$, batch size $B$, iteration number $T$, learning rate \learningrate[], momentum $p$, regularization coefficient \weightdecay{}.}
  \KwOut{final hypergradient of $\ztrain_i$ after $T$ iterations.}
  \tcc{initialization}
  $\paramdiff{0}{i}  \gets \vect{0}$ \\
  $\momdiff{0}{i} \gets \vect{0}$ \\
  \tcc{training}
  \For{$t \gets 1$ \KwTo{} $T$}{
    sample the current batch from the training dataset \\
    \eIf{the current batch contains $\ztrain_i$}
    { $\momdiff{t}{i} \gets p \momdiff{t-1}{i}  +  \weightdecay \paramdiff{t-1}{i}  + \frac{N}{B} \vect{g}_{t-1,i} $
    }
    {
      $\momdiff{t}{i} \gets p \momdiff{t-1}{i}  + \weightdecay \paramdiff{t-1}{i} $
    }
    $ \paramdiff{t}{i}\gets \paramdiff{t-1}{i} - \learningrate[t] \momdiff{t}{i} $ \\
  }

  \KwRet{ $\paramdiff{T}{i}$  }
  \caption{Hypergradient computation}
  \label{main-algorithm}
\end{algorithm}

\subsection{Fast Approximation}
We propose a fast approximation technique, which sets the Hessian \hessianer[t-1] in equation~\eqref{eq:sgd-hypergradient} and equation~\eqref{eq:momentum_diff_1} to zero with an analytical bound on the approximation error. In the data contribution equation~\eqref{eq:val_contrib}, the term $\pdv{\testloss( \param[T])}/{\param}$ can always be computed exactly, so this approximation only affects the $\paramdiff{T}{i}$ term. Algorithm~\ref{main-algorithm} shows the mini-batch version of the approximate algorithm for computing $\paramdiff{T}{i}$.

The Hessian-vector product can be approximated in \(\bigo(\norm{\param})\) time using a method akin to finite difference~\citep{Pearlmutter1994}. Nevertheless, the computation is slow as it requires (usually two) additional backpropagations and is susceptible to truncation and discretization errors just like finite difference. Consequently, removing the Hessian provides desirable simplification to the algorithm.


With the moderate conditions below, we can show that applying the proposed approximation results in bounded difference between the true $\paramdiff{t}{i}$ and its approximation $\appoxparamdiff{t}{i}$. Additionally, if we let the learning rate decay exponentially, the difference vanishes after sufficient training iterations.

\begin{condition}
  The training loss \(\trainingloss{}\) is twice differentiable.
\end{condition}

\begin{condition}
  The optimization process converges, that is,
  \begin{equation*}
    \lim_{t \to \infty}\param[t] = \optimalparam.
  \end{equation*}
\end{condition}

\begin{condition}
  \label{asmp:lipschitz}
  The empirical risk function $\empiricalrisk$ has Lipschitz-continuous gradients with Lipschitz constant $L$. Formally, there exists a constant $L$ such that
  \begin{equation*}
    \bignorm{\pdv{\empiricalrisk(\param[1])}{\param} - \pdv{\empiricalrisk(\param[2])}{\param}}_{2} \le L \norm{\param[1] - \param[2]}, \forall \param[1], \param[2] \in \hypothesisspace.
  \end{equation*}
\end{condition}

\begin{condition}
  The sequence of learning rates \learningrate{} is non-increasing and lower-bounded by 0. That is,
  \begin{equation*}
    \learningrate[t] \ge \learningrate[t+1] > 0, \forall t.
  \end{equation*}
\end{condition}

\begin{condition}
  The sequence of products of learning rate $\learningrate$ and regularization coefficient $\weightdecay$ satisfies
  \(0 < \learningrate[t]\weightdecay <1, \forall{t}\).
\end{condition}

\begin{condition}
  The sequence $\paramdiff{t}{i}$ is ultimately bounded by a constant $P$.
\end{condition}

\begin{theorem}\label{theorem:bounded_error}
  Under the above conditions and vanilla GD, the norm of the approximation error is bounded by \begin{equation*}
    \norm{\paramdiff{t}{i} - \appoxparamdiff{t}{i}} < L P \frac{ \learningrate[1]}{\learningrate[t]\weightdecay}.
  \end{equation*}
\end{theorem}

\begin{theorem} \label{theorem:vanishing_error}
  Under the above conditions, vanilla GD, and an exponential decay schedule for the learning rate $\eta$, the approximation error diminishes when $t$ tends to infinity
  \begin{equation*}
    \lim_{t \to \infty} \norm{\paramdiff{t}{i} - \appoxparamdiff{t}{i}} = 0.
  \end{equation*}
\end{theorem}
We can also relax the Lipschitz-continuity~\cref{asmp:lipschitz} and derive similar results. The proof details are in the supplemental material. In the experiments, we empirically verified these theoretical results and found that the approximation indeed leads to small errors.

\subsection{Time and Space Complexity}
It takes \(\bigo(\norm{\param})\) time and space to compute the parameter gradients. Using the approach in \citep{Pearlmutter1994}, Hessian-vector products can be computed in \(\bigo(\norm{\param})\) time and space.
We also need \(\bigo(\norm{\param})\) extra space to store previous \paramdiff{t-1}{i}, $\odv{\mom[t-1]}/{\epsilon_i}$ and $\vect{g}_{t-1,i}$.
Therefore, we need a total of \(\bigo(T\norm{\param})\) time and \(\bigo(\norm{\param})\)  space to trace a single training sample.
Tracing all training samples simultaneously requires \(\bigo(T\norm{\param}\abs{\trainingdataset})\) time and \(\bigo(\norm{\param}\abs{\trainingdataset})\) space.

\section{Related Work}
\subsubsection{Influence function.}
As a precursor to our work, \citep{koh2017understanding} measured the influence of a training data point $\ztrain$ on a test data point $\ztest$ via the obtained parameters $\param[T]$ as $\funname{IF}(\ztrain, \ztest)$
\begin{equation}
  \label{eq:IF}
  \funname{IF}(\ztrain, \ztest) \defeq - \pdv{\lossfun(\ztest, \param[T])}{\param}^{\mathsf{T}} \hessian[T]^{-1} \pdv{\lossfun(\ztrain, \param[T])}{\param}.
\end{equation}
The difficulty of equation~\eqref{eq:IF} lies in computing the inverse of Hessian $\hessian[T]^{-1}$, for which the naive approach is infeasible due to the size of modern DNNs. Koh \& Liang introduced two methods for this purpose.

The first method assumes that $\hessian$ is positive definite, $\hessian^{-1} \vect{v}$ can then be formulated as the optimal vector ${\vect{\phi}}^{*}$ that minimizes the quadratic form $\transpose{\vect{\phi}} \hessian \vect{\phi} + \transpose{\vect{v}}\vect{\phi}$ using conjugate gradient.

The second method to find $\hessian^{-1} \vect{v}$ is to iterate the following until convergence,
\begin{equation}
  \hessian^{-1} \vect{v} \leftarrow \vect{v} + \left(\matrx{I} - \pdv{\lossfun(\vect{z}, \param_T)}{\param} \right) \hessian^{-1} \vect{v},
  \label{eq:neumann-series}
\end{equation}
where the training data point $\vect{z}$ is randomly sampled in each iteration and the initial $\hessian^{-1} \vect{v}$ is set to $\vect{v}$.
In practice, we use the Monte Carlo expectation from multiple runs.
Both methods adopt the approximate Hessian-vector product~\citep{Pearlmutter1994}.

Although these techniques render the time and space complexity manageable, dealing with the inverse Hessian is still tedious and slow. Numerical considerations such as damping and scaling factors require careful treatment. Errors can be introduced by the variance of the Monte Carlo expectation and the errors in the Hessian-vector product. In comparison, \sysname{} offers a simplified method that removes the Hessian and a theoretical upper bound for the introduced error.


\subsubsection{Interpreting DNNs.}
A DNN prediction can be interpreted by the training data that have the most influence on the prediction. The work of \citet{Cook1980} on linear regression was one of the earliest techniques along this line of work.~\citep{sharchilev2018:tree-ensemble} studied tree ensembles.~\citep{koh2019accuracy} extended influence functions to the effects of groups of data points.~\citep{chen2020multistage} investigated the influence of data used in the pre-training on the fine-tuning task. Noting that outliers and mislabeled data often have outsize influence, \citep{barshan2020relatif} proposed RelatIF to accurately measure local influence. Other data-based interpretations utilize kernel functions~\citep{yeh2018representer,khanna2018fisher-kernels}, Shapley values~\citep{jia2019efficient,jia2019empirical,ghorbani2019data}, and network layers that identify prototypes and object parts~\citep{BransonVBP14,Been2016:criticism,QuanshiZhang2018,chen2019looks}.

\begin{figure*}[!t]
  \begin{subfigure}{0.33\textwidth}
    \includegraphics[scale=0.3]{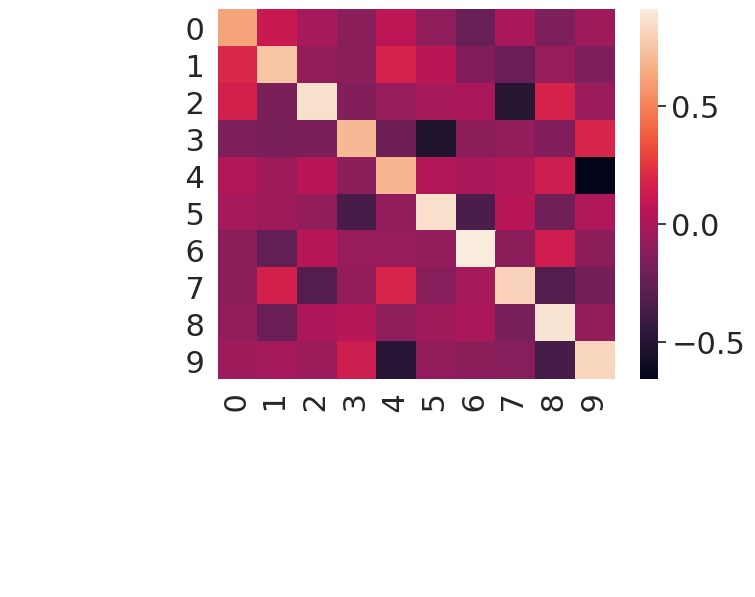}
    \caption{MNIST}
  \end{subfigure}
  \begin{subfigure}{0.33\textwidth}
    \includegraphics[scale=0.3]{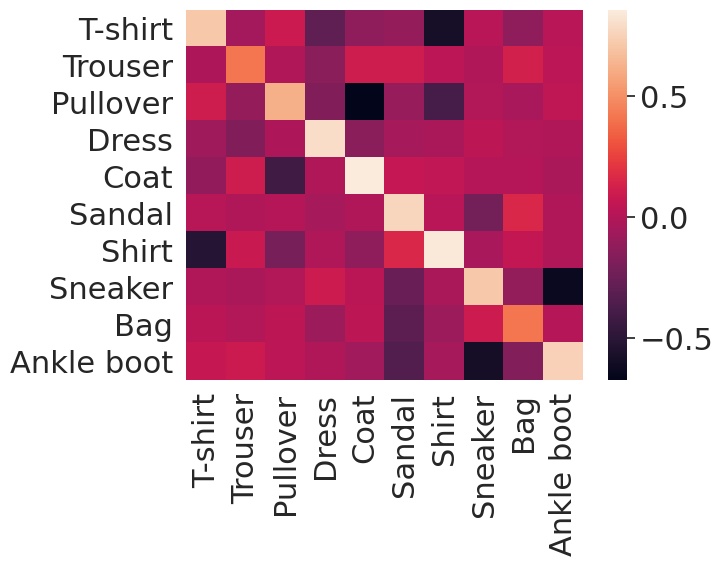}
    \caption{Fashion-MNIST}
  \end{subfigure}
  \begin{subfigure}{0.33\textwidth}
    \includegraphics[scale=0.3]{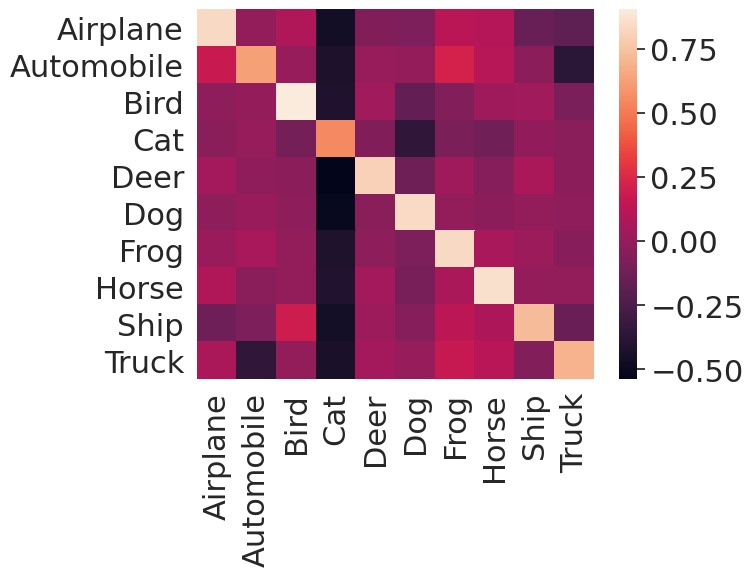}
    \caption{CIFAR-10}
  \end{subfigure}
  \caption{Inter-class contribution shown as heatmaps. Rows represent classes of training data and columns represent classes of test data.}\label{fig:heat_maps_of_three_cases}
\end{figure*}

Similar to our work, \citep{Satoshi2019} estimated the change of model parameters after the removal of a point by extrapolating via the Hessians along the training trajectory. In comparison, \sysname{} directly estimates the data contribution instead of the model parameters and disregards the unwieldy Hessian, which \citep{Satoshi2019} crucially depend on. Several works~\citep{litany2019soseleto,yoon2019data,ShuJun2019:Meta-Weight-Net} optimized the weights of individual training data points. The weights can serve as data valuation based on the entire test dataset, but cannot directly explain individual model predictions.

Neural networks can be interpreted from other perspectives. The weights of the network can be visualized~\citep{Zeiler2014,bau2018gandissect,fong2018}; behaviors of a complex network can be approximated with simple models that are easy to understand~\citep{lime2016,zhou2018:tree-stability,lakkaraju2017interpretable,Chen2019:distilling-into-concepts,ahern2019normlime}. In addition, model predictions can be explained by identifying important features in the input~\citep{simonyan2013deep,guided_backprop,smoothgrad,selvaraju2016gradcam,integrated_gradients,du2018:guided-feature-inversion}.

\subsubsection{Hypergradient-based optimization.}
Hypergradient is the gradient of the test loss with respect to the hyperparameters~\citep{Bengio1999:Hypergrad,domke2012,franceschi2017forward,franceschi2018bilevel,lorraine2019:millions}.
With hyperparameters $\vect{\epsilon}$, model parameters $\param$, and test loss $\testloss$, the hyperparameter optimization problem can be defined as the bi-level optimization
\begin{align*}
  \begin{split}
    \vect{\epsilon}^{*}                     & = \argmin_{\vect{\epsilon}} \testloss(\vect{\epsilon}, \param^{*}(\vect{\epsilon})), \\
    \subjectto \param^{*} (\vect{\epsilon}) & = \argmin_{\param} \trainingloss(\vect{\epsilon}, \param).
  \end{split}
\end{align*}
Note that we write $\param^{*} (\vect{\epsilon})$ to underscore the fact that $\vect{\epsilon}$ influences the optimization of $ \param^{*}$. Thus, in order to compute the gradient of $\validationloss(\vect{\epsilon}, \param^{*}(\vect{\epsilon}))$ w.r.t. $\vect{\epsilon}$, we must find the derivative $\pdv{\param^{*}(\vect{\epsilon})}/{\vect{\epsilon}}$, which may be computed by unrolling $\param^{*}$ throughout the training trajectory or for only a few steps as an approximation.

Hypergradient enables gradient-based optimization of hyperparameters, such as learning rates~\citep{donini2019marthe,metz2019understanding}, network architectures~\citep{darts2019}, or data augmentation~\citep{lin2019online}.
\citep{Maclaurin2015} and \citep{wang2018:dataset-distillation} distilled a large training dataset into \num{10}-\num{100} data points.~\citep{bohdal2020flexible} distilled data labels instead of input features.
\citep{mehra2019penalty} used penalty functions to avoid hypergradient in the bi-level optimization.

A key difference between \sysname{} and other works on hypergradient-based optimization is that we use the hypergradient $\paramdiff{t}{i}$ to assign credit to training data rather than to optimize $\vect{\epsilon}$. As a result, \sysname{} is less vulnerable to the accumulated inaccuracies in the hypergradient estimates over time. We take advantage of this by omitting the Hessian with bounded error.


\section{Experimental Evaluation}

\subsection{Datasets and Networks}
We used three image recognition datasets in our experiments: MNIST~\citep{Lecun1998}, Fashion-MNIST~\citep{xiao2017fashionmnistnovelimagedataset}, and CIFAR-10~\citep{Krizhevsky09learningmultiple}.
We chose two networks, LeNet-5~\citep{Lecun1998} of \num{61706} trainable parameters and DenseNet-40~\citep{huang2017densely} of \num{176122} trainable parameters. Details of the datasets and networks are discussed in the supplemental material.

\begin{table}[t!]
  \centering
  \caption{Training samples with extreme influence on the test data, their ground-truth and predicted labels, model confidence, and contribution on the test set. A positive contribution means the training sample reduces overall test loss}

  \begin{tabular}{@{} >{\centering\arraybackslash}m{4em}m{6em}m{5em}m{6em} @{}}
    \toprule
    {Training sample}                                                                                                                                                     & {True and predicted label}  & {Model confidence} & {Contribution to test data} \\
    \midrule
    \includegraphics{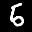}      & \centering{5 / 5}           & \centering{0.66}   & \num{-1.10e-4}              \\
    \includegraphics{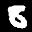}       & \centering{8 / 8}           & \centering{0.73}   & \num{-4.80e-5}              \\
    \includegraphics{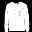} & \centering{Coat / Pullover} & \centering{0.66}   & \num{-9.39e-3}              \\
    \includegraphics{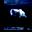}   & \centering{Deer / Deer}     & \centering{0.97}   & \num{-1.01e-5}              \\
  \end{tabular}
  \label{table:influential_samples}

\end{table}

\subsection{Manual Inspection}

In the first experiment, we verified that \sysname{} provides results that agree with our intuition. We computed the mean and standard deviation of the contributions of all training data points on $\testdataset$, which yields (\num{-3.17e-9}, \num{1.19e-6}) for MNIST, (\num{-4.9858e-7}, \num{0.1e-3}) for Fashion-MNIST, and (\num{-1.8676e-7}, \num{3.206e-6}) for CIFAR-10. The mean values were very close to zero, and the standard deviations were substantially greater than the mean. This is consistent with our expectation as the contribution of single data point in a large dataset is likely rather small. The standard deviations indicate that some data points have extreme contribution values, which we examine below.

\subsubsection{Influential examples.}
\Cref{table:influential_samples} shows data points from the three datasets with extreme contributions. The first two rows show examples from MNIST. The third and fourth rows are examples from Fashion-MNIST and CIFAR-10, respectively.
The first image is labeled as~\emph{5} but closely resembles~\emph{6}. The second is labeled as~\emph{8} and has an unusual upper half. Their average contribution to all test data points is four to five orders of magnitudes higher than the average ($\approx$ \num{e-9}). Due to their unusual appearances, these data points stand out from the rest of the training data and hence exert large influence on the model. Similarly, the two other examples have atypical appearances in their category and have large influences on the networks' predictions. Due to space considerations, we leave more examples to the supplemental material.


\subsubsection{Inter-class contribution.}
In the next experiment, we inspected the contribution from the training data of one class to the test data of another class. We partitioned the training dataset into $C$ subsets $S_{1}, \ldots, S_{C}$ according to the annotated class, where $C$ is the number of classes. Similarly, we partitioned the test set into $C$ subsets $T_{1}, \ldots, T_{C}$. The contribution $\mathcal{C}(S_{i}, T_{j})$ from a training class $S_{i}$ to a test class $T_{j}$ is the average contribution over all possible pairs of samples
\begin{equation*}
  \mathcal{C}(S_{i}, T_{j}) \defeq  \frac{1}{\abs{S_{i}}\abs{T_{j}}} \sum_{a \in S_{i} } \sum_{b \in T_{j}} \mathcal{C}(a,b) .
\end{equation*}

We plot the inter-class contribution matrices for MNIST, Fashion-MNIST, and CIFAR-10 in figure~\ref{fig:heat_maps_of_three_cases}. The rows represent training data classes and the columns represent test data classes.
As normalization, we divide every matrix entry by $\sqrt{\text{column sum} \times \text{row sum}}$.

In all three datasets, the maximum contribution values for each row and column appear on the diagonal, indicating that training data always produce the most reduction in test loss for their own class. In MNIST, we observe symmetrically low contribution between the digit pairs (3, 5), (2, 7), and (4, 9). As the two classes in each pair have similar appearances, there is competition between them. Lowering the test loss of one class will likely increase the test loss of the other class.
Similar symmetry exists for the pairs (Pullover, Coat), (T-shirt, Shirt), and (Sneaker, Ankle boot) in Fashion-MNIST, and between Automobile and Truck in CIFAR-10. We find these to be consistent with intuition.


\subsection{Approximation Error}
\begin{figure}[t]
  \centering
  \includegraphics[scale=0.55]{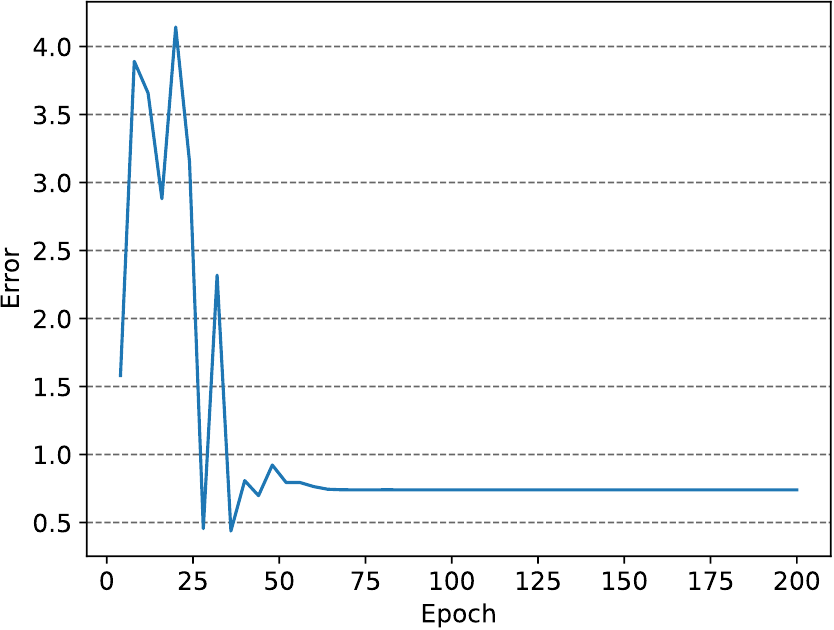}
  \caption{\sysname's hypergradient approximation error averaged over \num{500} Fashion-MNIST data points. }
  \label{fig:Fashion_MNIST_absolute_error}
\end{figure}

Theorems~\ref{theorem:bounded_error} and~\ref{theorem:vanishing_error} show that, under moderate conditions, the error from disregarding the Hessian term is bounded. In order to test this hypothesis, we tracked the contribution of \num{500} samples from Fashion-MNIST, including \num{1}\% from every class, through the training trajectory, using both the Hessian-aware method and the approximation of \sysname. We trained LeNet-5 for \num{200} epochs with the same hyperparameters as before and record the average $L^2$ norm of the approximation error $\norm{\paramdiff{t}{i} - \appoxparamdiff{t}{i}}$ every four epochs. Figure~\ref{fig:Fashion_MNIST_absolute_error} shows the results. We observed that the error stabilizes to \num{0.74} after about \num{70} epochs, which agrees with the theoretical result that the approximation error is bounded.

We further benchmarked \sysname's approximation method and influence functions~\citep{koh2017understanding} against the Hessian-aware method. While we recognized that it is difficult to establish any ground truth for data contribution values, both influence functions and \sysname{} may be considered as approximations of the Hessian-aware method. Therefore, we took the Hessian-aware data contribution values as the gold standard. In the experiments below, we computed the data contribution from influence functions using Monte Carlo expectation (equation~\ref{eq:neumann-series}) starting from epoch \num{100}, when the optimization is close to convergence.



\begin{figure}[t]
  \centering
  \includegraphics[scale=0.55]{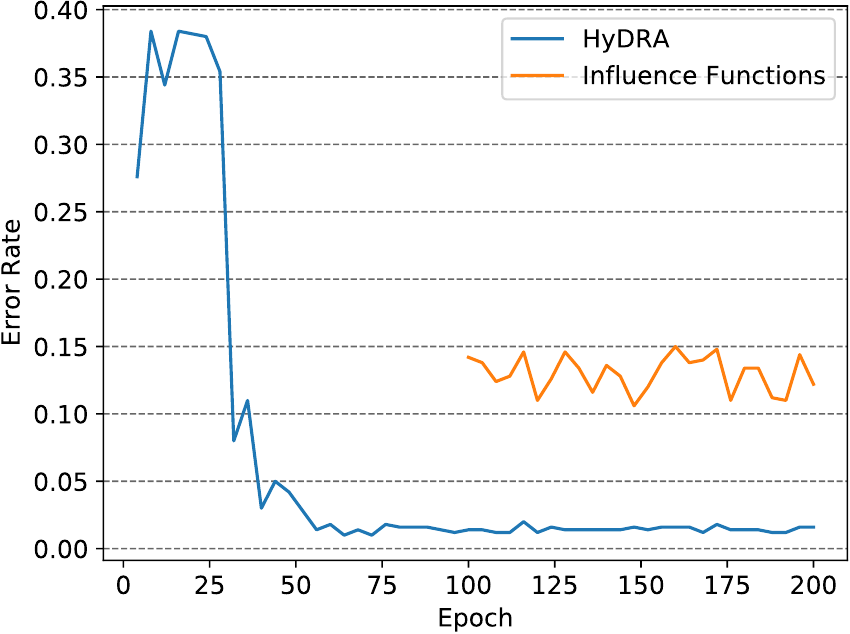}
  \caption{Sign error rates of \sysname's approximation and influence functions on Fashion-MNIST}\label{fig:Fashion_MNIST_contribution_sign_error_rate}
\end{figure}

\begin{figure}[t]
  \centering
  \includegraphics[scale=0.55]{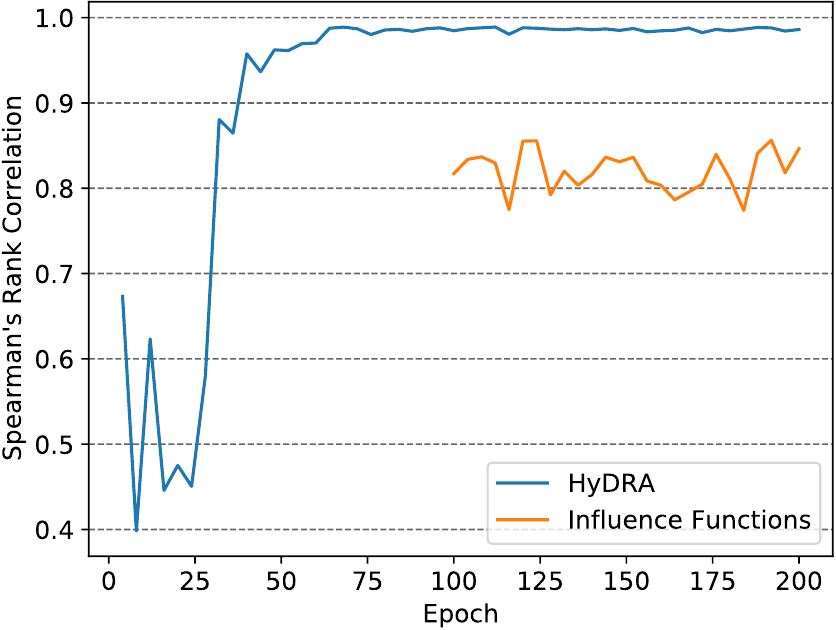}
  \caption{Spearman's rank correlation with the ground truth for \sysname's approximation technique and influence functions on Fashion-MNIST}\label{fig:Fashion_MNIST_spearman}
\end{figure}

We adopted two evaluation metrics used by \citep{koh2019accuracy}. First, we measured the percentage of data points where \sysname{} and influence functions erroneously flip the sign of the contribution (figure~\ref{fig:Fashion_MNIST_contribution_sign_error_rate}). That is, a data point making a positive contribution was mistakenly assigned negative contribution and vice versa. The error rate of \sysname{} converged after \num{100} epochs to an average of \num{0.014} with a standard deviation of \num{0.002}. The error rate of influence functions was \num{0.130} on average with a standard deviation of \num{0.013}.
Second, we ranked the data points by their contribution and compute Spearman's rank correlation against the Hessian-aware method (figure~\ref{fig:Fashion_MNIST_spearman}). After \num{100} epochs, the correlation of \sysname{} converged to \num{0.986} with a standard deviation of \num{0.002}, whereas influence functions showed substantially lower correlation with a mean of \num{0.820} that fluctuate more wildly (standard deviation = \num{0.024}).

In summary, the experimental comparison shows that \sysname's approximation to be not only more accurate but also an order of magnitude more stable than influence functions. This demonstrates that accounting for the entire trajectory is important in accurately estimating data influence. In these experiments, ignoring the optimization trajectory increases the error in the estimated contributions by \num{11.6}\% to \num{16.6}\%.

\subsection{Speedup by Approximation}
We found that removing the Hessian reduces the total running time by about \num{971} folds in a simple experiment. Specifically, tracking~\num{20000} data points on a DenseNet-40 network for \num{1} epoch using Hessian-vector products took about \num{49} hours on a server with two NVIDIA 2080Ti GPU cards, an AMD Ryzen 7 3800X 8-Core CPU, and \num{32} GB RAM. In contrast, the proposed fast approximation method reduced the training time to about \num{3} minutes.





\subsection{Debugging Training Data}
In this section, we investigate if \sysname{} can help debug datasets with erroneous labels.
In the first experiment (table~\ref{tab:remove-noisy-labels}), we compared the ability of different methods to clean noisy data. We created synthetic datasets with a known proportion of data points having randomly permuted, erroneous labels. We used \sysname{} and influence functions to estimate the data contribution and discard the least useful $r\%$ of training data. The network was retrained on the remaining training data. As DNNs are known to be robust against noise~\citep{rolnick2018deeplearningrobustmassive}, we adopted large error rates. The supplemental material contains detailed settings and additional experiments.
\sysname{} has demonstrated clear performance advantages over influence functions, which gradually increased as we went from the easy MNIST dataset to the most difficult CIFAR-10.

\begin{table}[t]
  \renewcommand{\arraystretch}{1.2}
  \centering
  \caption{Classification accuracy when different methods are used to clean the dataset with \num{80}\% proportion of label noise. }

  \begin{tabular}{@{} lcc @{}}
    \toprule
    {Dataset}                      & {Method}
                                   & {Final accuracy (\%)}                  \\


    \midrule
    \multirow{3}{*}{MNIST}         & No Filtering          & 72.93          \\
                                   & Influence Function    & 90.07          \\
                                   & \sysname{}            & \textbf{98.31} \\

    \midrule
    \multirow{3}{*}{Fashion-MNIST} & No Filtering          & 78.52          \\
                                   & Influence Function    & 63.20          \\
                                   & \sysname{}            & \textbf{86.72} \\
    \midrule
    \multirow{3}{*}{CIFAR-10}      & No Filtering          & 31.36          \\
                                   & Influence Function    & 42.12          \\
                                   & \sysname{}            & \textbf{72.01} \\
  \end{tabular}
  \label{tab:remove-noisy-labels}

\end{table}

In the second experiment, we aimed to detect data points with erroneous labels unsupervisedly by clustering the data points using their contribution to test data points. We created synthetic data by randomly selecting \num{50}\% of training data and uniformly assigning incorrect labels to them. After the models had been trained, for each training data point, we computed a $1000$-dimensional feature vector, which consisted of its contributions to \num{1000} randomly sampled validation samples. This feature vector was then discretized to $\{+1, -1\}^{1000}$ based on the signs of the values.

After that, we clustered the training data points using the discrete feature vectors, in the hope that the clustering can separate the correctly labeled from the incorrectly labeled. We performed the clustering separately for each class of training data points and calculate the average performance. As the performance metric, we calculated the Jaccard index, or the intersection over union, between the identified clusters and the ground-truth partition. Higher Jaccard indices indicate better clusters. We repeated this for every class label and reported the average.

\Cref{tab:clustering} reports the average Jaccard indices over \num{10} classes. We observe that Fashion-MNIST poses a much more difficult challenge to the unsupervised task than MNIST and causes both methods to obtain lower performance. On both datasets, \sysname{} outperforms influence functions; the performance gaps range from \num{25.25}\% to \num{44.51}\%. We attribute the superior performance of \sysname{} to the accuracy of the estimated data contribution values.

\begin{table}[t]
  \renewcommand{\arraystretch}{1.2}
  \caption{The overlap between automatically identified clusters of training data and the gold-standard clusters of correctly labeled and randomly labeled data}
  \centering

  \begin{tabular}{@{} lccc @{}}
    \toprule
    \multirow[c]{2}{*}[-0.5cm]{Dataset} & \multirow[c]{2}{*}[-0.5cm]{Method} & \multicolumn{2}{c}{Overlap (\%)}         \\
    \cmidrule{3-4}
                                        &                                    & \shortstack[t]{Correct                   \\ label} & \shortstack[t]{Random \\ label} \\
    \midrule
    \multirow{2}{*}{MNIST}              & \sysname{}                         & 97.13                            & 97.16 \\
                                        & Influence Function                 & 71.89                            & 67.19 \\
    \midrule
    \multirow{2}{*}{Fashion-MNIST}      & \sysname{}                         & 78.06                            & 77.73 \\
                                        & Influence Function                 & 41.26                            & 33.22 \\
  \end{tabular}
  \label{tab:clustering}

\end{table}


\section{Conclusion}
We propose \sysname{}, a technique for estimating the contribution of training data by differentiating the test loss against training data weights through time. To simplify computation, we provide a Hessian-free approximation to the exact derivative and establish an analytical upper bound for the approximation error. In the experiments, we compare \sysname{} to influence functions~\citep{koh2017understanding}, and confirm that \sysname{} provide more accurate estimates for data contribution, which facilitates the identification of label noise in the training data. With \sysname, the AI research community can be equipped with an effective and computationally efficient tool to interpret the influence of training samples to DNN predictions.

\section{Acknowledgments}
This research is supported by Alibaba Group through Alibaba Innovative Research (AIR) Program and Alibaba-NTU Singapore Joint Research Institute (JRI) (Alibaba-NTU-AIR2019B1), Nanyang Technological University, Singapore; the Nanyang Assistant Professorship (NAP); NTU-SDU-CFAIR (NSC-2019-011); the National Research Foundation, Singapore under its AI Singapore Programme (AISG Award No: AISG-GC-2019-003); the RIE 2020 Advanced Manufacturing and Engineering (AME) Programmatic Fund (No. A20G8b0102), Singapore; and the National Research Foundation, Singapore, Prime Minister's Office under its NRF Investigatorship Programme (NRFI Award No: NRF-NRFI05-2019-0002). Any opinions, findings, conclusions, or recommendations expressed in this material are those of the author(s) and do not reflect the views of the funding agencies.

\section{Broader Impact}
The deployment of machine learning in critical areas such as law enforcement and human resources has raised concerns regarding potential bias and prejudice of such algorithms~\citep{frazier2019learning,Poyiadzi2020:Face}. In many cases, the apparent bias is not due to algorithmic design, but to existing stereotypes inadvertently captured by training data~\citep{mehrabi2019survey,Bryant2019}. Therefore, the ability to trace an algorithmic prediction back to training data samples could help mitigate and eventually eliminate model bias in machine learning. We caution that the elimination of bias requires systemic efforts that extend far beyond pure algorithmic research. Having said that, we believe that this work could play a positive role in that direction.

\clearpage

\renewcommand{\appendixpagename}{Supplemental Material}
\appendix
\appendixpage

\input{supplemental_content}
\bibliography{hypergradient, ./bib_collection/paper/dataset, ./bib_collection/paper/machine_learning, ./bib_collection/paper/explainable_artificial_intellegence}

\end{document}

%% file: latex_definition/math.tex
\usepackage{amssymb,amsmath,amsfonts}
\usepackage{ifluatex}
\ifluatex
  \usepackage{unicode-math}
  \providecommand{\indicatorfun}{\mathbb{1}}
  \providecommand{\vect}[1]{\symbf{#1}}
  \providecommand{\matrx}[1]{\symbf{#1}}
  
\else
  \DeclareMathAlphabet{\mathpzc}{T1}{pzc}{m}{it}
  
  \providecommand{\indicatorfun}{\ensuremath{\mathbbm{1}}}
  \providecommand{\vect}[1]{\boldsymbol{#1}}
  \providecommand{\matrx}[1]{\boldsymbol{#1}}
  
\fi

\DeclareMathOperator*{\argmin}{arg\,min}

\providecommand{\bigo}{\ensuremath{O}}
\providecommand{\funname}[1]{\ensuremath{\operatorname{#1}}}
\providecommand{\abs}[1]{\ensuremath{\lvert#1\rvert}}
\providecommand{\ltwonorm}[1]{\ensuremath{{\lVert#1\rVert}_2}}
\providecommand{\bignorm}[1]{\ensuremath{{\Big \lVert#1 \Big\rVert}}}

\providecommand{\norm}[1]{\ltwonorm{#1}}

\providecommand{\defeq}{\ensuremath{\coloneq}}
\providecommand{\transpose}[1]{{#1}^{\mathsf{T}}}
\providecommand{\subjectto}{\ensuremath{\text{s.t. }}}

%% file: latex_definition/machinelearning.tex
\usepackage{xifthen}

\providecommand{\hypothesisspace}{\ensuremath{\mathcal{H}}}

\providecommand{\param}[1][]{\ensuremath{\ifthenelse{\isempty{#1}} {\vect{\theta}} {\vect{\theta}_{#1}}}}
\providecommand{\optimal}[1]{\ensuremath{{#1}^{*}}}
\providecommand{\optimalparam}[1][]{\ensuremath{\ifthenelse{\isempty{#1}} {\optimal{\param}} {\optimal{\param}_{\text{#1}}}}}
\providecommand{\finalparam}[1][]{\ensuremath{\ifthenelse{\isempty{#1}} {\hat{\vect{\theta}}} {\hat{\vect{\theta}}_{\text{#1}}}}}

\providecommand{\trainingdataset}[1][]{\ensuremath{
    \ifthenelse{\isempty{#1}}
    {D_{\text{train}}}
    {D_{\text{train,{#1}}}
    }}}
\providecommand{\validationdataset}[1][]{\ensuremath{
    \ifthenelse{\isempty{#1}}
    {D_{\text{val}}}
    {D_{\text{val,{#1}}}
    }}}
\providecommand{\batch}[1][]{\ensuremath{
    \ifthenelse{\isempty{#1}}
    {B}
    {B_{#1}
    }}}
\providecommand{\trainingdatasetsize}[1][]{\ensuremath{
    \ifthenelse{\isempty{#1}}
    {\abs{\trainingdataset}}
    {\abs{\trainingdataset[#1]}}
  }}
\providecommand{\testdatasetsize}[1][]{\ensuremath{
    \ifthenelse{\isempty{#1}}
    {\abs{\testdataset}}
    {\abs{\testdataset[#1]}}
  }}
\providecommand{\batchsize}[1][]{\ensuremath{
    \ifthenelse{\isempty{#1}}
    {\abs{\batch}}
    {\abs{\batch[#1]}}
  }}
\providecommand{\testdataset}{\ensuremath{D_{\text{test}}}}

\providecommand{\dataitem}{\ensuremath{\vect{z}}}
\providecommand{\trainingitem}[1][]{\ensuremath{\ifthenelse{\isempty{#1}} {\dataitem_\text{train}} {\dataitem_{#1}}}}
\providecommand{\testitem}[1][]{\ensuremath{\ifthenelse{\isempty{#1}} {\dataitem_\text{test}} {\dataitem_{#1}}}}
\providecommand{\sampleweight}[1][]{\ensuremath{\ifthenelse{\isempty{#1}} {\epsilon} {\epsilon_{#1}}}}

\providecommand{\mlfun}[1][]{\ensuremath{\ifthenelse{\isempty{#1}} {f_{\param}} {f_{#1}} }}
\providecommand{\weightdecay}{\ensuremath{\lambda}}
\providecommand{\learningrate}[1][]{\ensuremath{\ifthenelse{\isempty{#1}} {\eta} {\eta_{#1}}}}

\providecommand{\empiricalrisk}[1][]{\ensuremath{\ifthenelse{\isempty{#1}} {\mathcal{R}} {\mathcal{R}_\text{#1}}}}

\providecommand{\fungradient}[2]{\ensuremath{ \nabla_{#2} {#1}}}
\providecommand{\ergradient}[1][]{\ensuremath{ \ifthenelse{\isempty{#1}} {\fungradient{\empiricalrisk}{\param}} {\fungradient{\empiricalrisk}{#1}}}}

\providecommand{\mom}[1][]{\ensuremath{\ifthenelse{\isempty{#1}} {\vect{v}} {\vect{v}_{#1}}}}

\providecommand{\trainingloss}[1][]{\ensuremath{\ifthenelse{\isempty{#1}}
{\mathcal{L}_{\text{train}}}
{\mathcal{L}_{\text{#1}}}
}
}
\providecommand{\validationloss}[1][]{\ensuremath{\ifthenelse{\isempty{#1}}
{\mathcal{L}_{\text{val}}}
{\mathcal{L}_{\text{#1}}}
}
}
\providecommand{\batchloss}[1][]{\ensuremath{\ifthenelse{\isempty{#1}}
{\mathcal{L}_{\batch}}
{\mathcal{L}_{\batch[#1]}}
}}

\providecommand{\batchempiricalrisk}[1][]{\ensuremath{\ifthenelse{\isempty{#1}}
{\mathcal{R}_{\batch}}
{\mathcal{R}_{\batch[#1]}}
}}
\providecommand{\traininggradient}[1][]{\ensuremath{ \ifthenelse{\isempty{#1}} {\fungradient{\trainingloss}{\param}} {\fungradient{\trainingloss}{#1}}}}

\providecommand{\funhessian}[2]{\ensuremath{ \nabla^2_{#2} {#1}}}
\providecommand{\traininglosshessian}[1][]{\ensuremath{ \ifthenelse{\isempty{#1}} {\funhessian{\trainingloss}{\param}} {\funhessian{\trainingloss}{#1}}}}
\providecommand{\empiricalriskhessian}[1][]{\ensuremath{ \ifthenelse{\isempty{#1}} {\funhessian{\empiricalrisk}{\param}} {\funhessian{\empiricalrisk}{\param_{#1}}}}}

\providecommand{\regularizer}{\ensuremath{f_{\text{reg}}}}
\providecommand{\ltworegularizer}[1][]{\ensuremath{\frac{1}{2} \transpose{\param[#1]}\param[#1]}}
\providecommand{\lossfun}{\ensuremath{\ell}}

\providecommand{\testloss}{\ensuremath{\mathcal{L}_{\text{test}}}}

\providecommand{\ith}{\ensuremath{i{\text{th}}}}
\providecommand{\jth}{\ensuremath{j{\text{th}}}}
\providecommand{\tth}{\ensuremath{t{\text{th}}}}

%% file: latex_definition/theorem.tex
\declaretheorem[style=definition]{condition}
\declaretheorem{theorem}
\declaretheorem[style=definition]{definition}
\declaretheorem{corollary}

%% file: definitions.tex
\crefname{condition}{condition}{conditions}
\crefname{definition}{definition}{definitions}
\crefname{corollary}{corollary}{corollaries}

\newcommand{\ztest}[1][]{\ensuremath{\ifthenelse{\isempty{#1}}   {\vect{z}_{\text{test}}     } {\vect{z}_{\text{test},#1}}}}
\renewcommand{\batch}[1][]{\ensuremath{\ifthenelse{\isempty{#1}} {\text{batch}} {\text{batch}_{#1}} }}
\newcommand{\lossbatch}[1]{\ensuremath{\mathcal{L}_{\text{batch}_{#1}}}}
\newcommand{\errorv}[1][]{\ensuremath{\ifthenelse{\isempty{#1}} {\vect{e}} {\vect{e}_{#1}}}}

\renewcommand{\mom}[1][]{\ensuremath{\ifthenelse{\isempty{#1}} {\vect{v}} {\vect{v}_{#1}}}}
\newcommand{\paramdiff}[2]{\ensuremath{\nabla_{#1, #2}}}
\newcommand{\appoxparamdiff}[2]{\ensuremath{\widetilde{\nabla}_{#1, #2}}}
\newcommand{\momdiff}[2]{\ensuremath{\odv{\mom[#1]}{\epsilon_{#2}}}}
\newcommand{\persampleloss}[2]{\ensuremath{\lossfun(\vect{z}_{#1}, \param_{#2})}}
\newcommand{\maxeigenvalue}[1][]{\ensuremath{ \ifthenelse{\isempty{#1}} {\kappa_{\text{max}}} {\kappa_{\text{max}}(#1)}}}
\newcommand{\hessian}[1][]{\ensuremath{ \ifthenelse{\isempty{#1}} {\matrx{H}} {\matrx{H}_{#1}}}}
\newcommand{\hessianer}[1][]{\ensuremath{ \ifthenelse{\isempty{#1}} {\matrx{H}_{\text{er}}} {\matrx{H}_{\text{er},{#1}}}}}
\newcommand{\ztrain}{\ensuremath{\vect{z}}}

%% file: supplemental_content.tex
\setcounter{theorem}{0}
\setcounter{condition}{0}
\section{Experiment Settings}

MNIST and Fashion-MNIST were divided into training sets of \num{50000} samples, validation sets of \num{10000} samples, and test sets of \num{10000} samples, respectively. CIFAR-10 was divided into a training set of \num{40000} samples, a validation set of \num{10000} samples, and a test set of \num{10000} samples, respectively.


LeNet-5 consists of \num{61706} trainable parameters. Tanh activations were replaced with ReLU activations in our experiments. We also used the DenseNet-40 model with a growth rate of 12, which contains \num{176122} trainable parameters.

\Cref{table:hyperparameters} gives the detailed hyperparameters and other settings used in our experiments. For the reduce-on-plateau schedule, we recognize a plateau when the sum of the training loss and the validation loss does not decrease more than \num{0.01}\% of the best value found in two epochs.

All networks were optimized using SGD with momentum of \num{0.9}. On MNIST and Fashion-MNIST, we trained LeNet-5 using SGD with momentum for \num{20} epochs with a batch size of \num{64}. This procedure obtains test accuracies of \num{99.09}\% and \num{89.99}\% respectively.
On CIFAR-10, we trained DenseNet-40 using SGD with momentum for \num{150} epochs with a batch size of 64 and obtains test accuracy of \num{90.5}\%.

In the first training data debugging experiment, we stopped early after \num{20} epochs of training and report the accuracy in the last epoch. In the second training data debugging experiment, we trained the networks for \num{50} epochs.

\section{Additional Examples}
Two data points from MNIST with extreme contributions are shown in~\cref{table:MNIST_samples2}. The first is labeled as~\emph{5} but closely resembles~\emph{6}. The second is labeled as~\emph{8} and has an unusual upper half. Their average contribution to all test data points is four to five orders of magnitudes higher than the average ($\approx$ \num{e-9}). Due to their unusual appearances, these data points stand out from the rest of the training data and hence exert large influence on the model.

The two training samples heavily influence the two test data points in~\cref{table:MNIST_samples_influenced2}.
The first test sample in~\cref{table:MNIST_samples_influenced2} is misclassified as~\emph{5}.~\sysname{} indicates this error may be caused by the training sample that conflates~\emph{5} and~\emph{6}. The second test sample is wrongly classified as~\emph{3}. Regardless, the outlandish looking~\emph{8} probably helped in bringing the training loss on this digit down.


\begin{table}[t!]
  \centering
  \caption{MNIST training samples with extreme influence on the test data, their ground-truth and predicted labels, model confidence, and contribution on the test set. A positive contribution means the training sample reduces overall test loss}\label{table:MNIST_samples2}
  \sisetup{
    table-number-alignment = center,
  }
  \begin{tabular}{lcSS}
    \toprule
    \shortstack[t]{Training                                                                                                                                                                               \\ sample}                                                                                                                                                & \shortstack[t]{True and                               \\  predicted label} & \shortstack[t]{Model \\ confidence} & \shortstack[t]{Contribution \\ to test data} \\
    \midrule
    \adjustimage{valign=m}{hypergradient_analysis_result/MNIST/index_25678_contribution_-0.0001098600696423091_predicted_class_5_prob_0.6650401949882507_real_class_5.jpg} & 5 / 5 & 0.66 & \num{-1.1e-4} \\
    \vspace{.5\baselineskip}                                                                                                                                                                              \\
    \adjustimage{valign=m}{hypergradient_analysis_result/MNIST/index_4692_contribution_4.8405003326479346e-05_predicted_class_8_prob_0.7390213012695312_real_class_8.jpg}  & 8 / 8 & 0.73 & \num{-4.8e-5} \\
  \end{tabular}
\end{table}

\begin{table}[t!]
  \centering
  \caption{Test samples strongly influenced by training samples in~\cref{table:MNIST_samples2}}\label{table:MNIST_samples_influenced2}
  \begin{tabular}{@{} >{\centering\arraybackslash}m{4em} >{\centering\arraybackslash}m{4em}Scm{4em}@{}}
    \toprule
    {Test sample}                                                                                                                                                                    & {Influencer}                                                                                                                                                           & {Contrib.} & \shortstack[t]{True and        \\ predicted label}
                                                                                                                                                                                     & {Model confidence}                                                                                                                                                                                                   \\
    \midrule
    \adjustimage{valign=m}{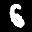} & \adjustimage{valign=m}{hypergradient_analysis_result/MNIST/index_25678_contribution_-0.0001098600696423091_predicted_class_5_prob_0.6650401949882507_real_class_5.jpg} & -0.319     & 6 / 5                   & 0.78 \\
    \vspace{.5\baselineskip}                                                                                                                                                                                                                                                                                                                                                                                \\
    \adjustimage{valign=m}{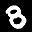}   & \adjustimage{valign=m}{hypergradient_analysis_result/MNIST/index_4692_contribution_4.8405003326479346e-05_predicted_class_8_prob_0.7390213012695312_real_class_8.jpg}  & 0.091      & 8 / 3                   & 0.80 \\
  \end{tabular}
\end{table}

\begin{table*}[h]
  \caption{Hyperparameters of experiments}\label{table:hyperparameters}

  \centering
  \sisetup{
    table-number-alignment = center,
  }
  \begin{tabular}{lSSSccS}
    \toprule
    \thead{Dataset} & {\thead{Epoch                                            \\ number}} & {\thead{Batch     \\ size}} & {\thead{{Initial} \\ {learning rate}}} & \thead{Learning rate \\ schedule} & \thead{Weight decay} & {\thead{Momentum}} \\
    \midrule
    MNIST           & 20            & 64  & 0.01 & \makecell{Multiplied by 0.1 \\ at epoch 5
    }               & \num{2.0e-5}  & 0.9                                      \\
    \vspace{.5\baselineskip}                                                   \\
    Fashion-MNIST   & 20            & 64  & 0.01 & \makecell{Multiplied by 0.5 \\ on plateaus of 2 epochs} & \num{2.0e-5}  & 0.9                    \\
    \vspace{.5\baselineskip}                                                   \\
    CIFAR-10        & 150           & 64  & 0.1  & \makecell{Multiplied by 0.1 \\ on plateaus of 2 epochs} & \num{2.5e-5}  & 0.9                    \\
  \end{tabular}
\end{table*}

\begin{table}[t]
  \centering
  \caption{Classification accuracy when different methods are used to clean the dataset with \num{50}\% proportion of label noise}
  \label{tab:remove-noisy-labels2}
  \begin{tabular}{@{} ccc @{}}
    \toprule
    {Dataset}                      & {Method}
                                   & {Final                     accuracy (\%)}                  \\
    \midrule
    \multirow{3}{*}{MNIST}         & No Filtering                              & 94.65          \\
                                   & Influence Function                        & 93.83          \\
                                   & \sysname{}                                & \textbf{98.31} \\
    \midrule
    \multirow{3}{*}{Fashion-MNIST} & No Filtering                              & \textbf{87.44} \\
                                   & Influence Function                        & 76.70          \\
                                   & \sysname{}                                & 87.28          \\
    \midrule
    \multirow{3}{*}{CIFAR-10}      & No Filtering                              & 68.91          \\
                                   & Influence Function                        & 58.34          \\
                                   & \sysname{}                                & \textbf{75.06} \\
  \end{tabular}
\end{table}



\section{Additional Results for Debugging Training Data}
We report the results for debugging training data with the label noise rate set to \num{50}\% in table~\ref{tab:remove-noisy-labels2}.
For all datasets, training the network from scratch using samples chosen by \sysname{} leads to better accuracy than influence functions. In addition, \sysname{} produces significantly better performance than training directly on noisy data on MNIST and CIFAR10. An exception happens on Fashion-MNIST, where filtering noisy data points using either method performs worse than not filtering the data at all.

\section{Growth of Training Time}
Our theoretical analysis indicates that the training time of \sysname{} scales linearly with with the number of model parameters and the number of data points whose contributions are tracked through the training trajectory. In this section, we empirically study how training time of \sysname{} grows with those two factors.


We performed the experiments on a server with an AMD Ryzen 7 3800X 8-Core processor, \num{32} GB of main memory, two GeForce RTX 2080 Ti GPU cards, each with 12GB memory, and a 1 TB XPG GAMMIX S50 solid state drive. We used three different networks, LeNet5~\citep{Lecun1998}, DenseNet-40~\citep{huang2017densely}, and MobileNet V2~\citep{sandler2019mobilenetv2}, which have \num{61706}, \num{176122}, and \num{2236682} trainable parameters respectively. For each network, we recorded the training time when tracking the contribution of \num{400}, \num{2000} and \num{10000} training data points.

\begin{figure}[t]
  \centering
  \includegraphics[scale=0.55]{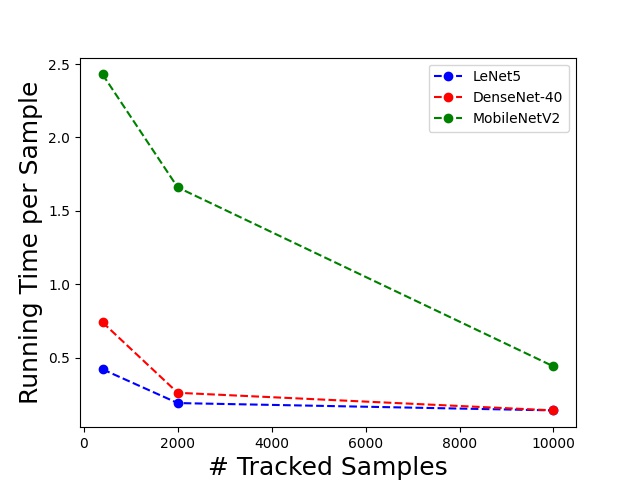}
  \includegraphics[scale=0.55]{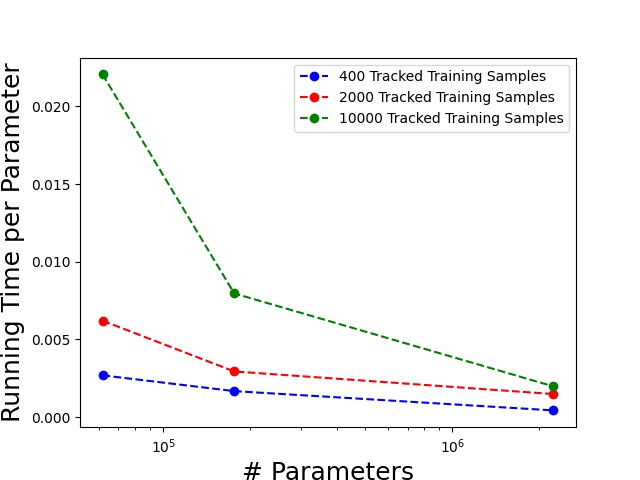}
  \caption{The training time scales up with respect to \#Parameters and \#Tracked Samples.}
  \label{fig:scale_up}
\end{figure}

\Cref{fig:scale_up} shows the average training time per tracked data point and per network parameter. We note that the average training time does not increase, which indicates the training time scales sublinearly initially and grows linearly afterwards.

\section{Approximation Error Analysis}\label{appendix:approximation_error_analysis}
In this section, we will analyze the approximation error introduced by dropping the Hessian term in the case of vanilla GD\@.
After dropping the Hessian term, the recurrent update equation becomes $\appoxparamdiff{t}{i}  = \paramdiff{t-1}{i} -\learningrate \weightdecay \paramdiff{t-1}{i} - \learningrate \vect{g}_{t-1,i}$.
We are interested in bounding the norm of the approximation error, which is defined below.

\begin{definition}
  The approximation error at the \tth{} iteration is defined as
  $\errorv[t]  \defeq \paramdiff{t}{i} - \appoxparamdiff{t}{i}$,
  where $\appoxparamdiff{t}{i}$ is the approximation of $\paramdiff{t}{i}$.
\end{definition}

Before the analysis, we also need some moderate conditions about the optimization process.

\begin{condition}\label{asmp:differentiable}
  The training loss \(\trainingloss{}\) is twice differentiable.
\end{condition}

\begin{condition}
  The optimization process converges, that is, $\lim_{t \to \infty}\param[t] = \optimalparam$.
\end{condition}

\subsection{Bounds with Lipschitz Continuity}

First, we bound the error under several moderate conditions regarding the optimization process. In the next subsection, we show how these conditions can be further relaxed.

\begin{condition}\label{asmp:lipschitz-gradient}
  The empirical risk function $\empiricalrisk$ has Lipschitz-continuous gradients with Lipschitz constant $L$. Formally, there exists a constant $L$ such that
  \begin{equation*}
    \bignorm{\pdv{\empiricalrisk(\param[1])}{\param} - \pdv{\empiricalrisk(\param[2])}{\param}}_{2} \le L \norm{\param[1] - \param[2]}, \forall \param[1], \param[2] \in \hypothesisspace.
  \end{equation*}
\end{condition}
\noindent The Lipschitz continuity is a regular condition, which is implied by, for example, that $\empiricalrisk$ is twice differentiable and $\param$ takes value from a compact hypothesis space. The latter is likely true since $\optimalparam$ tends to be close to $\param[0]$. It is worth noting that this condition constrains the eigenvalues of $\hessianer[t]$ to the range $[-L, L]$.

\begin{condition}
  The learning rate sequence \learningrate{} is non-increasing and lower-bounded by 0. That is,
  \begin{equation*}
    \learningrate[t] \ge \learningrate[t+1] > 0, \forall t.
  \end{equation*}
\end{condition}

Since $\learningrate[t]$ and $\weightdecay$ are both typically quite small, we assume their product is also small.
\begin{condition}
  The sequence of products of learning rate $\learningrate$ and regularization coefficient $\weightdecay$ satisfies
  \(0 < \learningrate[t]\weightdecay <1, \forall{t}\).
\end{condition}

Finally, the sequence $\paramdiff{t}{i}$ should be bounded, or efforts to estimate it would end in vain.
\begin{condition}\label{asmp:limit-bound-contrib}
  The sequence $\paramdiff{t}{i}$ is ultimately bounded by a constant $P$. Without loss of generality,
  \begin{equation*}
    \norm{\paramdiff{t}{i}} < P,  \forall i.
  \end{equation*}
\end{condition}

\begin{theorem}
  With conditions~\ref{asmp:differentiable}-\ref{asmp:limit-bound-contrib}, the norm of the approximation error is bounded by \begin{equation*}
    \norm{\errorv[t]} < L P \frac{ \learningrate[1]}{\learningrate[t]\weightdecay}.
  \end{equation*}
\end{theorem}
\noindent Proof.

First, we have the recursive formula:
\begin{align}
  \errorv[0] & =\vect{0},                                                                                     \\
  \errorv[t] & =(1-\learningrate \weightdecay)\errorv[t-1]- \learningrate \hessianer[t-1] \paramdiff{t-1}{i}.
\end{align}
After solving it, we get
\begin{align}
  \errorv[t] & = \sum_{j=1}^{t} (-\learningrate[j]){(1-\learningrate[j] \weightdecay)}^{t-j} \hessianer[j-1]\paramdiff{j-1}{i}.
\end{align}
By the triangle inequality,
\begin{equation*}
  \norm{\errorv[t]} \leq
  \sum_{j=1}^{t} \learningrate[j]{(1-\learningrate[j] \weightdecay)}^{t-j} \norm{\hessianer[j-1]\paramdiff{j-1}{i}}.
\end{equation*}
We then have
\begin{equation*}
  \begin{split}
         & \sum_{j=1}^{t} \learningrate[j]{(1-\learningrate[j] \weightdecay)}^{t-j} \norm{\hessianer[j-1]\paramdiff{j-1}{i}} \\
    \leq & \, \sum_{j=1}^{t} \learningrate[j]{(1-\learningrate[j] \weightdecay)}^{t-j} \norm{L \paramdiff{j-1}{i}}           \\
    =    & \, \sum_{j=1}^{t} \learningrate[j]{(1-\learningrate[j] \weightdecay)}^{t-j} L P                                   \\
    \leq & \,  L P \learningrate[1] \sum_{j=0}^{t-1}{(1-\learningrate[t] \weightdecay)}^{j}                                  \\
    =    & \,  L P \learningrate[1] \frac{1-(1-\learningrate[t]\weightdecay)^{t}}{\learningrate[t]\weightdecay}              \\
    <    & \,  L P \frac{ \learningrate[1]}{\learningrate[t]\weightdecay}.
  \end{split}
\end{equation*}
As such, we obtain the desired inequality.~\qed{}

Furthermore, if we know that the learning rate decays sufficiently exponentially, the approximation error diminishes when $t$ tends to infinity.
\begin{condition}\label{asmp:lr-decay}
  The learning rate sequence \learningrate{} decays exponentially at rate $c$, which is less than $1-\learningrate[1]\weightdecay$. That is,
  \begin{equation*}
    \learningrate[t+1] = c  \learningrate[t], \; \forall t ,
  \end{equation*}
  \begin{equation*}
    0 < c < 1-\learningrate[1]\weightdecay.
  \end{equation*}
\end{condition}

\begin{theorem}
  With conditions~\ref{asmp:differentiable}-\ref{asmp:lr-decay} and the learning rate schedule in condition~\ref{asmp:lr-decay}, the approximation error diminishes when $t$ tends to infinity
  \begin{equation*}
    \lim_{t \to \infty} \norm{\errorv[t]} = 0.
  \end{equation*}
\end{theorem}

\noindent Proof.

Under the specific learning rate schedule, we have
\begin{equation*}
  \begin{split}
    \norm{\errorv[t]}
     & \leq L P  \sum_{j=1}^{t}\learningrate[j] {(1-\learningrate[j] \weightdecay)}^{t-j}                                                                                                                                          \\
     & = L P  \sum_{j=1}^{t}\learningrate[1] c^{j-1} {(1-\learningrate[j] \weightdecay)}^{t-j}                                                                                                                                     \\
     & \leq L P \frac{\learningrate[1]}{c} {(1-\learningrate[t] \weightdecay)}^t \sum_{j=1}^{t}\left(\frac{c}{1-\learningrate[1] \weightdecay}\right)^{j}                                                                          \\
     & = L P \frac{\learningrate[1]}{c} {(1-\learningrate[t] \weightdecay)}^t \frac{\frac{c}{1-\learningrate[1] \weightdecay} - (\frac{c}{1-\learningrate[1] \weightdecay})^{t+1}}{1 - \frac{c}{1-\learningrate[1] \weightdecay}}.
  \end{split}
\end{equation*}
When $t \to \infty$, ${(1-\learningrate[t] \weightdecay)}^t$ and $(c / (1-\learningrate[1] \weightdecay))^{t+1}$ go to zero, and the claim follows.~\qed{}

\subsection{Bounds with Relaxed Conditions}

The condition~\ref{asmp:lipschitz-gradient} in the previous section may appear to be too restrictive. In this subsection, we replace it with a more relaxed condition and show that the error is still upper-bounded in the limit.



\begin{condition}\label{asmp:continuous_hessian}
  The Hessian sequence \hessianer{} converges as $t \to \infty$
  \begin{align}
    \lim_{t \to \infty} \hessianer[t] =\optimal{\hessianer},
  \end{align}
  where $\optimal{\hessianer}$ is the Hessian of the empirical risk $\empiricalrisk$ at $\optimalparam$.
\end{condition}
Since $\param$ converges, it makes sense to assume that Hessian converges, too.

\begin{corollary}
  The eigenvalues of \hessianer{} converge to the eigenvalues of $\optimal{\hessianer}$.
\end{corollary}

By this corollary, we can find an index $N$ such that $\abs{\maxeigenvalue{\hessianer[t]} -\maxeigenvalue{\optimal{\hessianer}}} < \delta, \forall t \geq N $, given an arbitrarily small $\delta$, where \(\maxeigenvalue\) is the eigenvalue with maximal absolute value. Now, we fix such an infinitesimal $\delta_{\text{eigen}}$ and the corresponding index $N_{\text{eigen}}$.


Finally, since $\learningrate[t]$ and $\weightdecay$ are both typically quite small, we assume their product is eventually small.

\begin{theorem}
  With conditions~\ref{asmp:differentiable}-\ref{asmp:limit-bound-contrib}, and~\ref{asmp:continuous_hessian}, and let $t_0 = N_{\text{eigen}} +1 $ be the start of the tail portion of the optimization, we can upper bound the error's norm as
  \begin{equation*}
    \lim_{t \to \infty} \norm{\errorv[t, t_0]} < (\maxeigenvalue{\optimal{\hessianer}}+\delta_{\text{eigen}}) P \frac{ \learningrate[t_0]}{\learningrate[t]\weightdecay}.
  \end{equation*}
  where $\errorv[t, t_0]$ is a shorthand for a two-part sum that constitute $\errorv[t]$,
  \begin{equation*}
    \begin{split}
      \errorv[t] & = \errorv[t, t_0]                                                                                                       \\
                 & = \sum_{j=1}^{t_0-1} (-\learningrate[j]){(1-\learningrate[j] \weightdecay)}^{t-j} \hessianer[j-1]\paramdiff{j-1}{i}     \\
                 & \quad +\sum_{j=t_0}^{t} (-\learningrate[j]){(1-\learningrate[j] \weightdecay)}^{t-j} \hessianer[j-1]\paramdiff{j-1}{i}.
    \end{split}
  \end{equation*}
\end{theorem}

Proof.
Again, we have the recursive formula:
\begin{align}
  \errorv[0] & =\vect{0},                                                                                     \\
  \errorv[t] & =(1-\learningrate \weightdecay)\errorv[t-1]- \learningrate \hessianer[t-1] \paramdiff{t-1}{i}.
\end{align}
After solving it, we get
\begin{align}
  \errorv[t] & = \sum_{j=1}^{t} (-\learningrate[j]){(1-\learningrate[j] \weightdecay)}^{t-j} \hessianer[j-1]\paramdiff{j-1}{i}.
\end{align}
Now if \(t \geq t_0\), then by the triangle inequality,
\begin{equation*}
  \begin{split}
    \norm{\errorv[t, t_0]} & \leq
    \sum_{j=1}^{t_0-1} \learningrate[j]{(1-\learningrate[j] \weightdecay)}^{t-j} \norm{\hessianer[j-1]\paramdiff{j-1}{i}}                                \\
                           & \quad +\sum_{j=t_0}^{t} \learningrate[j]{(1-\learningrate[j] \weightdecay)}^{t-j} \norm{\hessianer[j-1]\paramdiff{j-1}{i}}.
  \end{split}
\end{equation*}

Consider the second part first, we have
\begin{equation*}
  \begin{split}
         & \sum_{j=t_0}^{t} \learningrate[j]{(1-\learningrate[j] \weightdecay)}^{t-j} \norm{\hessianer[j-1]\paramdiff{j-1}{i}}                                                 \\
    <    & \sum_{j=t_0}^{t} \learningrate[j]{(1-\learningrate[j] \weightdecay)}^{t-j} (\maxeigenvalue{\optimal{\hessianer}}+\delta_{\text{eigen}}) \norm{\paramdiff{j-1}{i}}   \\
    \leq & (\maxeigenvalue{\optimal{\hessianer}}+\delta_{\text{eigen}}) P \sum_{j=t_0}^{t} \learningrate[j]{(1-\learningrate[j] \weightdecay)}^{t-j}                           \\
    \leq & (\maxeigenvalue{\optimal{\hessianer}}+\delta_{\text{eigen}}) P \learningrate[t_0] \sum_{j=t_0}^{t}{(1-\learningrate[t] \weightdecay)}^{t-j}                         \\
    =    & (\maxeigenvalue{\optimal{\hessianer}}+\delta_{\text{eigen}}) P \learningrate[t_0] \frac{1-(1-\learningrate[t]\weightdecay){}^{t-t_0}}{\learningrate[t]\weightdecay} \\
    <    & (\maxeigenvalue{\optimal{\hessianer}}+\delta_{\text{eigen}}) P \frac{ \learningrate[t_0]}{\learningrate[t]\weightdecay}.
  \end{split}
\end{equation*}

Note that the first part of the right-hand side goes to $0$ as $t \to \infty$. In other words, for  any small $\delta_1 > 0$, there exists a number $N_1 > t_0$, such that for all $t \geq N_1$,
\begin{equation*}
  \sum_{j=1}^{t_0-1} \learningrate[j]{(1-\learningrate[j] \weightdecay)}^{t-j} \norm{\hessianer[j-1]\paramdiff{j-1}{i}} < \delta_1.
\end{equation*}

Taken together, for any infinitesimal $\delta_1 > 0$, there exist an index $N_1$ such that
\begin{equation*}
  \norm{\errorv[t, t_0]} < (\maxeigenvalue{\optimal{\hessianer}}+\delta_{\text{eigen}}) P \frac{ \learningrate[t_0]}{\learningrate[t]\weightdecay} + \delta_1, \forall t \geq N_1.
\end{equation*}
This is the definition of the limit that we seek to prove.~\qed{}

If we add condition~\ref{asmp:lr-decay}, we would have the same conclusion as before.
\begin{theorem}
  With the extra condition~\ref{asmp:lr-decay},
  \begin{equation*}
    \lim_{t \to \infty}
    \norm{\errorv[t, t_0]} = 0.
  \end{equation*}
\end{theorem}
\noindent Proof.

Reconsidering the second part, which can be rewritten as
\begin{equation*}
  \begin{split}
         & \sum_{j=t_0}^{t} \learningrate[j]{(1-\learningrate[j] \weightdecay)}^{t-j} \norm{\hessianer[j-1]\paramdiff{j-1}{i}}                                                                                                      \\
    \leq & (\maxeigenvalue{\optimal{\hessianer}}+\delta_{\text{eigen}}) P \sum_{j=t_0}^{t} \learningrate[j]{(1-\learningrate[j] \weightdecay)}^{t-j}                                                                                \\
    \leq & (\maxeigenvalue{\optimal{\hessianer}}+\delta_{\text{eigen}}) P \sum_{j=t_0}^{t} \learningrate[t_0]c^{j-t_0}{(1-\learningrate[j] \weightdecay)}^{t-j}                                                                     \\
    =    & (\maxeigenvalue{\optimal{\hessianer}}+\delta_{\text{eigen}}) P \frac{\learningrate[L]}{c^{t_0}} \sum_{j=t_0}^{t} c^{j}{(1-\learningrate[j] \weightdecay)}^{t-j}                                                          \\
    \leq & (\maxeigenvalue{\optimal{\hessianer}}+\delta_{\text{eigen}}) P \frac{\learningrate[t_0]}{c^{t_0}}  {(1-\learningrate[t] \weightdecay)}^t    \sum_{j=t_0}^{t} \left(\frac{c}{1-\learningrate[j] \weightdecay}\right)^{j}  \\
    \leq & (\maxeigenvalue{\optimal{\hessianer}}+\delta_{\text{eigen}}) P \frac{\learningrate[t_0]}{c^{t_0}}  {(1-\learningrate[t] \weightdecay)}^t    \sum_{j=t_0}^{t} \left(\frac{c}{1-\learningrate[1] \weightdecay}\right)^{j}.
  \end{split}
\end{equation*}
Introducing $q= \frac{c}{1-\learningrate[1] \weightdecay}$ for simplification, we have
\begin{equation*}
  \begin{split}
         & \sum_{j=t_0}^{t} \learningrate[j]{(1-\learningrate[j] \weightdecay)}^{t-j} \norm{\hessianer[j-1]\paramdiff{j-1}{i}}                                                    \\
    \leq & (\maxeigenvalue{\optimal{\hessianer}}+\delta_{\text{eigen}}) P \frac{\learningrate[t_0]}{c^{t_0}}  {(1-\learningrate[t] \weightdecay)}^t   \frac{q^{t_0}-q^{t+1}}{1-q} \\
    \leq & (\maxeigenvalue{\optimal{\hessianer}}+\delta_{\text{eigen}}) P \frac{\learningrate[t_0]}{c^{t_0}}  {(1-\learningrate[t] \weightdecay)}^t   \frac{q^{t_0}}{1-q}.
  \end{split}
\end{equation*}
Finally, since $\lim_{t \to \infty} {(1-\learningrate[t] \weightdecay)}^t=0$, the desired conclusion follows.~\qed{}

\section{Mini-Batch Hypergradient}\label{appendix:minibatch_derivation}
Here we consider mini-batch training with a batch size of \(B\), and other symbols are of the same meanings as above.
Formally, the loss function at \tth{} batch is
\begin{equation*}
  \begin{split}
    \lossbatch{t}( \param[t-1]) & = \frac{1}{B} \sum_{ \vect{z} \in \batch[t]} \, \persampleloss{}{t-1}          \\
                                & \quad + \indicatorfun_{batch_{t}}(i)*\frac{N\epsilon}{B}\persampleloss{i}{t-1} \\
                                & \quad +\ltworegularizer[t-1] ,
  \end{split}
\end{equation*}
where the indicator function $\indicatorfun_{batch_{t}}$ is introduced to determine whether the traced training point $\vect{z}_i$ is in $batch_{t}$.

As before, we have the initial conditions
\begin{align}
  \paramdiff{0}{i} & = \vect{0}, \\
  \momdiff{0}{i}   & = \vect{0}.
\end{align}

And the recurrence formula becomes
\begin{align*}
  \odv{\mom[t]}{\epsilon_i} & = p \odv{\mom[t-1]}{\epsilon_i} + \hessianer[t-1] \paramdiff{t-1}{i}                    \\
                            & \quad + \weightdecay \paramdiff{t-1}{i}                                                 \\
                            & \quad + \indicatorfun_{batch_{t}}(i)* \pdv{\persampleloss{i}{t-1}}{\param}*\frac{N}{B}, \\
  \paramdiff{t}{i}          & = \paramdiff{t-1}{i} - \learningrate \odv{\mom[t]} {\epsilon_i},
\end{align*}
where \hessianer[t] denotes the Hessian of the regularizer-free batch loss.
Also, we could omit \hessianer[t] here.

%% file: main.bbl
\begin{thebibliography}{60}
\providecommand{\natexlab}[1]{#1}
\providecommand{\url}[1]{\texttt{#1}}
\providecommand{\urlprefix}{URL }
\expandafter\ifx\csname urlstyle\endcsname\relax
  \providecommand{\doi}[1]{doi:\discretionary{}{}{}#1}\else
  \providecommand{\doi}{doi:\discretionary{}{}{}\begingroup \urlstyle{rm}\Url}\fi

\bibitem[{Ahern et~al.(2019)Ahern, Noack, Guzman-Nateras, Dou, Li, and Huan}]{ahern2019normlime}
Ahern, I.; Noack, A.; Guzman-Nateras, L.; Dou, D.; Li, B.; and Huan, J. 2019.
\newblock NormLime: A New Feature Importance Metric for Explaining Deep Neural Networks.
\newblock \emph{arXiv Preprint 1909.04200} .

\bibitem[{Alaa and Van Der~Schaar(2020)}]{pmlr-v119-alaa20b}
Alaa, A.; and Van Der~Schaar, M. 2020.
\newblock Frequentist Uncertainty in Recurrent Neural Networks via Blockwise Influence Functions.
\newblock In III, H.~D.; and Singh, A., eds., \emph{Proceedings of the 37th International Conference on Machine Learning}, volume 119 of \emph{Proceedings of Machine Learning Research}, 175--190. PMLR.
\newblock \urlprefix\url{https://proceedings.mlr.press/v119/alaa20b.html}.

\bibitem[{Arora, Cohen, and Hazan(2018)}]{arora2018optimization}
Arora, S.; Cohen, N.; and Hazan, E. 2018.
\newblock On the Optimization of Deep Networks: Implicit Acceleration by Overparameterization.
\newblock \emph{arXiv 1802.06509} .

\bibitem[{Barshan, Brunet, and Dziugaite(2020)}]{barshan2020relatif}
Barshan, E.; Brunet, M.-E.; and Dziugaite, G.~K. 2020.
\newblock Relatif: Identifying explanatory training samples via relative influence.
\newblock In \emph{International Conference on Artificial Intelligence and Statistics}, 1899--1909. PMLR.

\bibitem[{Bau et~al.(2018)Bau, Zhu, Strobelt, Bolei, Tenenbaum, Freeman, and Torralba}]{bau2018gandissect}
Bau, D.; Zhu, J.-Y.; Strobelt, H.; Bolei, Z.; Tenenbaum, J.~B.; Freeman, W.~T.; and Torralba, A. 2018.
\newblock {GAN} Dissection: Visualizing and Understanding Generative Adversarial Networks.
\newblock \emph{arXiv preprint arXiv:1811.10597} .

\bibitem[{Bengio(1999)}]{Bengio1999:Hypergrad}
Bengio, Y. 1999.
\newblock Continuous Optimization of Hyper-Parameters.
\newblock Technical report, Department of Computer Science and Operations Research, University of Montreal.

\bibitem[{Bohdal, Yang, and Hospedales(2020)}]{bohdal2020flexible}
Bohdal, O.; Yang, Y.; and Hospedales, T. 2020.
\newblock Flexible Dataset Distillation: Learn Labels Instead of Images.
\newblock \emph{arXiv 2006.08572} .

\bibitem[{Branson et~al.(2014)Branson, Horn, Belongie, and Perona}]{BransonVBP14}
Branson, S.; Horn, G.~V.; Belongie, S.; and Perona, P. 2014.
\newblock Bird Species Categorization Using Pose Normalized Deep Convolutional Nets.
\newblock In \emph{BMVC}.

\bibitem[{Bryant and Howard(2019)}]{Bryant2019}
Bryant, D.; and Howard, A. 2019.
\newblock A Comparative Analysis of Emotion-Detecting AI Systems with Respect to Algorithm Performance and Dataset Diversity.
\newblock In \emph{Proceedings of the 2019 AAAI/ACM Conference on AI, Ethics, and Society}.

\bibitem[{Chen et~al.(2019)Chen, Li, Tao, Barnett, Su, and Rudin}]{chen2019looks}
Chen, C.; Li, O.; Tao, C.; Barnett, A.~J.; Su, J.; and Rudin, C. 2019.
\newblock This Looks Like That: Deep Learning for Interpretable Image Recognition.
\newblock In \emph{NeurIPS}.

\bibitem[{Chen et~al.(2020)Chen, Si, Li, Chelba, Kumar, Boning, and Hsieh}]{chen2020multistage}
Chen, H.; Si, S.; Li, Y.; Chelba, C.; Kumar, S.; Boning, D.; and Hsieh, C.-J. 2020.
\newblock Multi-Stage Influence Function.
\newblock \emph{arXiv 2007.09081} .

\bibitem[{{Chen} et~al.(2019){Chen}, {Chen}, {Huang}, {Ren}, and {Zhang}}]{Chen2019:distilling-into-concepts}
{Chen}, R.; {Chen}, H.; {Huang}, G.; {Ren}, J.; and {Zhang}, Q. 2019.
\newblock Explaining Neural Networks Semantically and Quantitatively.
\newblock In \emph{ICCV}.

\bibitem[{Cook and Weisberg(1980)}]{Cook1980}
Cook, R.~D.; and Weisberg, S. 1980.
\newblock Characterizations of an empirical influence function for detecting influential cases in regression.
\newblock \emph{Technometrics} 22(4): 495–508.

\bibitem[{Domke(2012)}]{domke2012}
Domke, J. 2012.
\newblock Generic Methods for Optimization-Based Modeling.
\newblock In \emph{AISTATS}.

\bibitem[{Donini et~al.(2020)Donini, Franceschi, Pontil, Majumder, and Frasconi}]{donini2019marthe}
Donini, M.; Franceschi, L.; Pontil, M.; Majumder, O.; and Frasconi, P. 2020.
\newblock MARTHE: Scheduling the Learning Rate Via Online Hypergradients.
\newblock In \emph{IJCAI}.

\bibitem[{Du et~al.(2018)Du, Liu, Song, and Hu}]{du2018:guided-feature-inversion}
Du, M.; Liu, N.; Song, Q.; and Hu, X. 2018.
\newblock Towards explanation of {DNN}-based prediction with guided feature inversion.
\newblock In \emph{KDD}.

\bibitem[{Fong and Vedaldi(2018)}]{fong2018}
Fong, R.; and Vedaldi, A. 2018.
\newblock Net2Vec: Quantifying and Explaining how Concepts are Encoded by Filters in Deep Neural Networks.
\newblock \emph{arXiv preprint arXiv:1801.03454} .

\bibitem[{Franceschi et~al.(2017)Franceschi, Donini, Frasconi, and Pontil}]{franceschi2017forward}
Franceschi, L.; Donini, M.; Frasconi, P.; and Pontil, M. 2017.
\newblock Forward and Reverse Gradient-Based Hyperparameter Optimization.
\newblock \emph{arXiv 1703.01785} .

\bibitem[{Franceschi et~al.(2018)Franceschi, Frasconi, Salzo, Grazzi, and Pontil}]{franceschi2018bilevel}
Franceschi, L.; Frasconi, P.; Salzo, S.; Grazzi, R.; and Pontil, M. 2018.
\newblock Bilevel Programming for Hyperparameter Optimization and Meta-Learning.
\newblock \emph{arXiv 1806.04910} .

\bibitem[{Frazier et~al.(2019)Frazier, Nahian, Riedl, and Harrison}]{frazier2019learning}
Frazier, S.; Nahian, M. S.~A.; Riedl, M.; and Harrison, B. 2019.
\newblock Learning Norms from Stories: A Prior for Value Aligned Agents.
\newblock In \emph{Proceedings of the AAAI/ACM Conference on AI, Ethics, and Society}.

\bibitem[{Ghorbani and Zou(2019)}]{ghorbani2019data}
Ghorbani, A.; and Zou, J. 2019.
\newblock Data Shapley: Equitable Valuation of Data for Machine Learning.
\newblock In \emph{International Conference on Machine Learning}, 2242--2251.

\bibitem[{Hara, Nitanda, and Maehara(2019)}]{Satoshi2019}
Hara, S.; Nitanda, A.; and Maehara, T. 2019.
\newblock Data Cleansing for Models Trained with SGD.
\newblock In \emph{NeurIPS}.

\bibitem[{Huang et~al.(2017)Huang, Liu, van~der Maaten, and Weinberger}]{huang2017densely}
Huang, G.; Liu, Z.; van~der Maaten, L.; and Weinberger, K.~Q. 2017.
\newblock Densely Connected Convolutional Networks.
\newblock In \emph{Proceedings of the IEEE Conference on Computer Vision and Pattern Recognition}.

\bibitem[{Jia et~al.(2019{\natexlab{a}})Jia, Dao, Wang, Hubis, Hynes, Gurel, Li, Zhang, Song, and Spanos}]{jia2019efficient}
Jia, R.; Dao, D.; Wang, B.; Hubis, F.~A.; Hynes, N.; Gurel, N.~M.; Li, B.; Zhang, C.; Song, D.; and Spanos, C. 2019{\natexlab{a}}.
\newblock Towards Efficient Data Valuation Based on the Shapley Value.
\newblock In \emph{AISTATS}.

\bibitem[{Jia et~al.(2019{\natexlab{b}})Jia, Sun, Xu, Zhang, Li, and Song}]{jia2019empirical}
Jia, R.; Sun, X.; Xu, J.; Zhang, C.; Li, B.; and Song, D. 2019{\natexlab{b}}.
\newblock An Empirical and Comparative Analysis of Data Valuation with Scalable Algorithms.
\newblock \emph{arXiv Preprint 1911.07128} .

\bibitem[{Khanna et~al.(2018)Khanna, Kim, Ghosh, and Koyejo}]{khanna2018fisher-kernels}
Khanna, R.; Kim, B.; Ghosh, J.; and Koyejo, O. 2018.
\newblock Interpreting Black Box Predictions using Fisher Kernels.
\newblock In \emph{ICML}.

\bibitem[{Kim, Khanna, and Koyejo(2016)}]{Been2016:criticism}
Kim, B.; Khanna, R.; and Koyejo, O.~O. 2016.
\newblock Examples are not enough, learn to criticize! Criticism for Interpretability.
\newblock In \emph{NeurIPS}.

\bibitem[{Koh et~al.(2019)Koh, Ang, Teo, and Liang}]{koh2019accuracy}
Koh, P.~W.; Ang, K.-S.; Teo, H. H.~K.; and Liang, P. 2019.
\newblock On the Accuracy of Influence Functions for Measuring Group Effects.
\newblock In \emph{NeurIPS}.

\bibitem[{Koh and Liang(2017)}]{koh2017understanding}
Koh, P.~W.; and Liang, P. 2017.
\newblock Understanding black-box predictions via influence functions.
\newblock In \emph{Proceedings of the 34th International Conference on Machine Learning}.

\bibitem[{Krizhevsky(2009)}]{Krizhevsky09learningmultiple}
Krizhevsky, A. 2009.
\newblock Learning multiple layers of features from tiny images.
\newblock Technical report, University of Toronto.

\bibitem[{Lakkaraju et~al.(2017)Lakkaraju, Kamar, Caruana, and Leskovec}]{lakkaraju2017interpretable}
Lakkaraju, H.; Kamar, E.; Caruana, R.; and Leskovec, J. 2017.
\newblock Interpretable \& explorable approximations of black box models.
\newblock \emph{arXiv preprint arXiv:1707.01154} .

\bibitem[{{Lecun} et~al.(1998){Lecun}, {Bottou}, {Bengio}, and {Haffner}}]{Lecun1998}
{Lecun}, Y.; {Bottou}, L.; {Bengio}, Y.; and {Haffner}, P. 1998.
\newblock Gradient-based learning applied to document recognition.
\newblock \emph{Proceedings of the IEEE} 86(11): 2278--2324.

\bibitem[{Li and Liang(2018)}]{NEURIPS2018_54fe976b}
Li, Y.; and Liang, Y. 2018.
\newblock Learning Overparameterized Neural Networks via Stochastic Gradient Descent on Structured Data.
\newblock In Bengio, S.; Wallach, H.; Larochelle, H.; Grauman, K.; Cesa-Bianchi, N.; and Garnett, R., eds., \emph{Advances in Neural Information Processing Systems}, volume~31. Curran Associates, Inc.
\newblock \urlprefix\url{https://proceedings.neurips.cc/paper_files/paper/2018/file/54fe976ba170c19ebae453679b362263-Paper.pdf}.

\bibitem[{Lin et~al.(2019)Lin, Guo, Li, Xin, Wu, Lin, Ouyang, and Yan}]{lin2019online}
Lin, C.; Guo, M.; Li, C.; Xin, Y.; Wu, W.; Lin, D.; Ouyang, W.; and Yan, J. 2019.
\newblock Online Hyper-parameter Learning for Auto-Augmentation Strategy.
\newblock \emph{arXiv 1905.07373} .

\bibitem[{Litany and Freedman(2019)}]{litany2019soseleto}
Litany, O.; and Freedman, D. 2019.
\newblock SOSELETO: A Unified approach to transfer learning and training with noisy labels.
\newblock In \emph{ICLR Workshop on Learning from Limited Labeled Data}.

\bibitem[{Liu, Simonyan, and Yang(2019)}]{darts2019}
Liu, H.; Simonyan, K.; and Yang, Y. 2019.
\newblock DARTS: Differentiable Architecture Search.
\newblock In \emph{ICLR}.

\bibitem[{Lorraine, Vicol, and Duvenaud(2020)}]{lorraine2019:millions}
Lorraine, J.; Vicol, P.; and Duvenaud, D. 2020.
\newblock Optimizing Millions of Hyperparameters by Implicit Differentiation.
\newblock In \emph{AISTATS}.

\bibitem[{Maclaurin, Duvenaud, and Adams(2015)}]{Maclaurin2015}
Maclaurin, D.; Duvenaud, D.; and Adams, R. 2015.
\newblock Gradient-based hyperparameter optimization through reversible learning.
\newblock In \emph{ICML}.

\bibitem[{Mehra and Hamm(2019)}]{mehra2019penalty}
Mehra, A.; and Hamm, J. 2019.
\newblock Penalty Method for Inversion-Free Deep Bilevel Optimization.
\newblock \emph{arXiv 1911.03432} .

\bibitem[{Mehrabi et~al.(2019)Mehrabi, Morstatter, Saxena, Lerman, and Galstyan}]{mehrabi2019survey}
Mehrabi, N.; Morstatter, F.; Saxena, N.; Lerman, K.; and Galstyan, A. 2019.
\newblock A Survey on Bias and Fairness in Machine Learning.
\newblock \emph{arXiv Preprint arXiv: 1908.09635} .

\bibitem[{Metz et~al.(2019)Metz, Maheswaranathan, Nixon, Freeman, and Sohl-Dickstein}]{metz2019understanding}
Metz, L.; Maheswaranathan, N.; Nixon, J.; Freeman, C.~D.; and Sohl-Dickstein, J. 2019.
\newblock Understanding and correcting pathologies in the training of learned optimizers.
\newblock In \emph{ICML}.

\bibitem[{Pearlmutter(1994)}]{Pearlmutter1994}
Pearlmutter, B.~A. 1994.
\newblock Fast Exact Multiplication by the Hessian.
\newblock \emph{Neural Computation} 6(1): 147--160.

\bibitem[{Poyiadzi et~al.(2020)Poyiadzi, Sokol, Santos-Rodriguez, De~Bie, and Flach}]{Poyiadzi2020:Face}
Poyiadzi, R.; Sokol, K.; Santos-Rodriguez, R.; De~Bie, T.; and Flach, P. 2020.
\newblock FACE: Feasible and Actionable Counterfactual Explanations.
\newblock In \emph{Proceedings of the AAAI/ACM Conference on AI, Ethics, and Society}.

\bibitem[{Ribeiro, Singh, and Guestrin(2016)}]{lime2016}
Ribeiro, M.~T.; Singh, S.; and Guestrin, C. 2016.
\newblock ``Why Should {I} Trust You?'': Explaining the Predictions of Any Classifier.
\newblock In \emph{KDD}.

\bibitem[{Rolnick et~al.(2018)Rolnick, Veit, Belongie, and Shavit}]{rolnick2018deeplearningrobustmassive}
Rolnick, D.; Veit, A.; Belongie, S.; and Shavit, N. 2018.
\newblock Deep Learning is Robust to Massive Label Noise.
\newblock \urlprefix\url{https://arxiv.org/abs/1705.10694}.

\bibitem[{Sandler et~al.(2019)Sandler, Howard, Zhu, Zhmoginov, and Chen}]{sandler2019mobilenetv2}
Sandler, M.; Howard, A.; Zhu, M.; Zhmoginov, A.; and Chen, L.-C. 2019.
\newblock MobileNetV2: Inverted Residuals and Linear Bottlenecks.
\newblock In \emph{CVPR}.

\bibitem[{Selvaraju et~al.(2017)Selvaraju, Cogswell, Das, Vedantam, Parikh, and Batra}]{selvaraju2016gradcam}
Selvaraju, R.~R.; Cogswell, M.; Das, A.; Vedantam, R.; Parikh, D.; and Batra, D. 2017.
\newblock Grad-{CAM}: Visual Explanations from Deep Networks via Gradient-based Localization.
\newblock \emph{ICCV} .

\bibitem[{Sharchilev et~al.(2018)Sharchilev, Ustinovsky, Serdyukov, and de~Rijke}]{sharchilev2018:tree-ensemble}
Sharchilev, B.; Ustinovsky, Y.; Serdyukov, P.; and de~Rijke, M. 2018.
\newblock Finding influential training samples for gradient boosted decision trees.
\newblock \emph{arXiv preprint arXiv:1802.06640} .

\bibitem[{Shu et~al.(2019)Shu, Xie, Yi, Zhao, Zhou, Xu, and Meng}]{ShuJun2019:Meta-Weight-Net}
Shu, J.; Xie, Q.; Yi, L.; Zhao, Q.; Zhou, S.; Xu, Z.; and Meng, D. 2019.
\newblock Meta-Weight-Net: Learning an explicit mapping for sample weighting.
\newblock In \emph{NeurIPS}.

\bibitem[{Simonyan, Vedaldi, and Zisserman(2013)}]{simonyan2013deep}
Simonyan, K.; Vedaldi, A.; and Zisserman, A. 2013.
\newblock Deep Inside Convolutional Networks: Visualising Image Classification Models and Saliency Maps.
\newblock \emph{arXiv 1312.6034} .

\bibitem[{Smilkov et~al.(2017)Smilkov, Thorat, Kim, Vi{\'{e}}gas, and Wattenberg}]{smoothgrad}
Smilkov, D.; Thorat, N.; Kim, B.; Vi{\'{e}}gas, F.~B.; and Wattenberg, M. 2017.
\newblock Smooth{G}rad: Removing noise by adding noise.
\newblock In \emph{ICML}.

\bibitem[{Springenberg et~al.(2014)Springenberg, Dosovitskiy, Brox, and Riedmiller}]{guided_backprop}
Springenberg, J.~T.; Dosovitskiy, A.; Brox, T.; and Riedmiller, M. 2014.
\newblock Striving for Simplicity: The All Convolutional Net.
\newblock In \emph{ICLR Workshop}.

\bibitem[{Sundararajan, Taly, and Yan(2017)}]{integrated_gradients}
Sundararajan, M.; Taly, A.; and Yan, Q. 2017.
\newblock Axiomatic Attribution for Deep Networks.
\newblock In \emph{Proceedings of the International Conference on Machine Learning}.

\bibitem[{Wang et~al.(2018)Wang, Zhu, Torralba, and Efros}]{wang2018:dataset-distillation}
Wang, T.; Zhu, J.-Y.; Torralba, A.; and Efros, A.~A. 2018.
\newblock Dataset Distillation.
\newblock \emph{arXiv 1811.10959} .

\bibitem[{Xiao, Rasul, and Vollgraf(2017)}]{xiao2017fashionmnistnovelimagedataset}
Xiao, H.; Rasul, K.; and Vollgraf, R. 2017.
\newblock Fashion-MNIST: a Novel Image Dataset for Benchmarking Machine Learning Algorithms.
\newblock \urlprefix\url{https://arxiv.org/abs/1708.07747}.

\bibitem[{Yeh et~al.(2018)Yeh, Kim, Yen, and Ravikumar}]{yeh2018representer}
Yeh, C.-K.; Kim, J.~S.; Yen, I. E.~H.; and Ravikumar, P. 2018.
\newblock Representer Point Selection for Explaining Deep Neural Networks.
\newblock In \emph{{NeurIPS}}.

\bibitem[{Yoon, Arik, and Pfister(2019)}]{yoon2019data}
Yoon, J.; Arik, S.~O.; and Pfister, T. 2019.
\newblock Data Valuation using Reinforcement Learning.
\newblock In \emph{ICML}.

\bibitem[{Zeiler and Fergus(2014)}]{Zeiler2014}
Zeiler, M.~D.; and Fergus, R. 2014.
\newblock Visualizing and Understanding Convolutional Networks.
\newblock In \emph{Proceedings of the European Conference on Computer Vision (ECCV)}, 818--833.

\bibitem[{Zhang, Wu, and Zhu(2018)}]{QuanshiZhang2018}
Zhang, Q.; Wu, Y.~N.; and Zhu, S.-C. 2018.
\newblock Interpretable Convolutional Neural Networks.
\newblock \emph{CVPR} .

\bibitem[{Zhou, Zhou, and Hooker(2018)}]{zhou2018:tree-stability}
Zhou, Y.; Zhou, Z.; and Hooker, G. 2018.
\newblock Approximation trees: Statistical stability in model distillation.
\newblock \emph{arXiv preprint arXiv:1808.07573} .

\end{thebibliography}
